\documentclass{article} 
\usepackage{preprint_style,times}

\usepackage{amsmath,amsfonts,bm}

\def\eqref#1{equation~\ref{#1}}
\def\Eqref#1{Equation~\ref{#1}}

\def\1{\bm{1}}

\DeclareMathAlphabet{\mathsfit}{\encodingdefault}{\sfdefault}{m}{sl}
\SetMathAlphabet{\mathsfit}{bold}{\encodingdefault}{\sfdefault}{bx}{n}

\usepackage{graphicx}
\usepackage{xcolor}
\usepackage{tikz}
\usepackage{flafter}
\usepackage{placeins}
\usepackage{amsthm}
\usepackage{amssymb}
\usepackage{mathtools}
\usepackage{booktabs}
\usepackage{multirow}
\usepackage{array}
\usepackage{float}
\usepackage{microtype}
\usepackage{enumitem}
\usepackage{mdframed}
\IfFileExists{algorithm.sty}{%
  \usepackage{algorithm}
}{%
  \floatstyle{ruled}
  \newfloat{algorithm}{tbp}{loa}
  \floatname{algorithm}{Algorithm}
}
\usetikzlibrary{arrows.meta,fit,positioning,calc,matrix,backgrounds}

\definecolor{latentblue}{HTML}{7E9DB5}
\definecolor{latentfill}{HTML}{ECF0F4}
\definecolor{selfgreen}{HTML}{8DAF88}
\definecolor{selffill}{HTML}{EEF3ED}
\definecolor{predictiongold}{HTML}{C4A87A}
\definecolor{predictionfill}{HTML}{F6F2EB}
\definecolor{fixedgray}{HTML}{6B7280}
\definecolor{fixedfill}{HTML}{F3F4F6}

\mdfdefinestyle{samplebox}{%
  linecolor=fixedgray!45,
  backgroundcolor=fixedfill,
  linewidth=0.4pt,
  leftmargin=0pt,
  rightmargin=0pt,
  innerleftmargin=7pt,
  innerrightmargin=7pt,
  innertopmargin=6pt,
  innerbottommargin=6pt,
  skipabove=6pt,
  skipbelow=8pt,
  splittopskip=6pt,
  splitbottomskip=6pt
}

\theoremstyle{remark}

\newcommand{\method}{Untied Self-Conditioning}
\newcommand{\methodshort}{Ours}
\newcommand{\papertitle}{Improving Few-Step Language Flows with\\ Untied Self-Conditioning}

\newcommand{\norm}[1]{\left\lVert #1 \right\rVert}
\newcommand{\ip}[2]{\left\langle #1,#2 \right\rangle}

\newcommand{\dd}{\mathrm{d}}

\newcommand{\RMS}{\operatorname{RMS}}

\newcommand{\diag}{\operatorname{Diag}}

\newcommand{\sg}{\operatorname{sg}}

\usepackage{hyperref}
\usepackage{url}
\hypersetup{
  hidelinks,
  pdftitle={Improving Few-Step Language Flows with Untied Self-Conditioning},
  pdfauthor={Bocheng Li and Linli Xu}
}

\title{\papertitle}

\author{%
  Bocheng Li$^{1,2}$ \qquad
  Linli Xu$^{1,2}$\thanks{Corresponding author.}\\
  \normalfont $^1$University of Science and Technology of China\\
  \normalfont $^2$State Key Laboratory of Cognitive Intelligence\\
  \normalfont\texttt{bcli@mail.ustc.edu.cn} \qquad
  \normalfont\texttt{linlixu@ustc.edu.cn}
}

\preprintcopy
\begin{document}

\maketitle

\begin{abstract}
Flow-matching language models refine all token positions in parallel and can
trade sampling steps for latency, yet generation quality still degrades sharply
with few sampling steps.  We trace a source of this degradation to a
train--inference mismatch in previous-prediction self-conditioning: during
training, the self-conditioning input is computed from the current noisy state
with no intervening solver step; during sampling, the solver folds the
previous prediction into the latent before that same prediction reappears as
the explicit self-conditioning input.  This coupling, absent during training,
creates redundancy that grows with step width.  We show that the mismatch
degrades both the self-conditioning input and the solver update, and derive a
correction for each from the model's own structure.  From the frozen projection weights we identify
directions along which the self-conditioning input is redundant with the
latent and dampen them; from the solver's integration structure we derive that
a step-average prediction is needed and approximate it from prediction history,
with scale set by offline trajectory statistics.  The resulting sampler,
\method{}, requires no retraining and uses one evaluation per step.  At 8
sampling steps on LangFlow, it reduces OpenWebText generative perplexity from
$531$ to~$62$ ($8.6\times$); under an adapted Arena-Hard-Auto~v2 protocol,
its outputs are preferred in $96\%$ of pairwise comparisons.  On ELF-B it
reduces generative perplexity from $71$ to~$43$.  Improvements hold from 8 to
256 sampling steps.
\end{abstract}

\section{Introduction}
\label{sec:introduction}

Flow matching~\citep{flow_matching} has enabled high-quality parallel
generation in continuous domains, with recent progress extending to
text~\citep{diffusion_lm,diffuseq,sed,plaid,flowseq,langflow,elf}.  These
models embed discrete tokens in a continuous latent space and refine all
positions simultaneously, approaching autoregressive quality while admitting
variable-cost sampling: fewer iterative steps directly reduce
latency~\citep{langflow,elf}.  Making each model evaluation as informative as
possible is the central challenge for few-step generation.

A standard technique in these models is previous-prediction
self-conditioning~\citep{analog_bits}: at each step, the model receives its own
clean prediction from the preceding step as additional
input~\citep{sed,tess,plaid,seqdiffuseq,fmseq,elf}.  Self-conditioning
consistently improves generation quality, yet it introduces a
train--inference mismatch: during sampling, the solver folds the previous
prediction into the latent before that same prediction reappears as the
explicit self-conditioning input, creating a coupling between the two input
pathways that is absent during training and grows with step width.
Prior work has studied train--inference discrepancies in the self-conditioning
recurrence~\citep{tencdm,trec,fastdiss} and proposed inference-time
modifications to individual components of the sampling
loop~\citep{ace,dpm_solver_pp,higs}, but has not diagnosed the specific
mechanism by which this coupling degrades the self-conditioning input or the
solver's own update.

\begin{figure}[!t]
\centering
\includegraphics[width=0.9\linewidth]{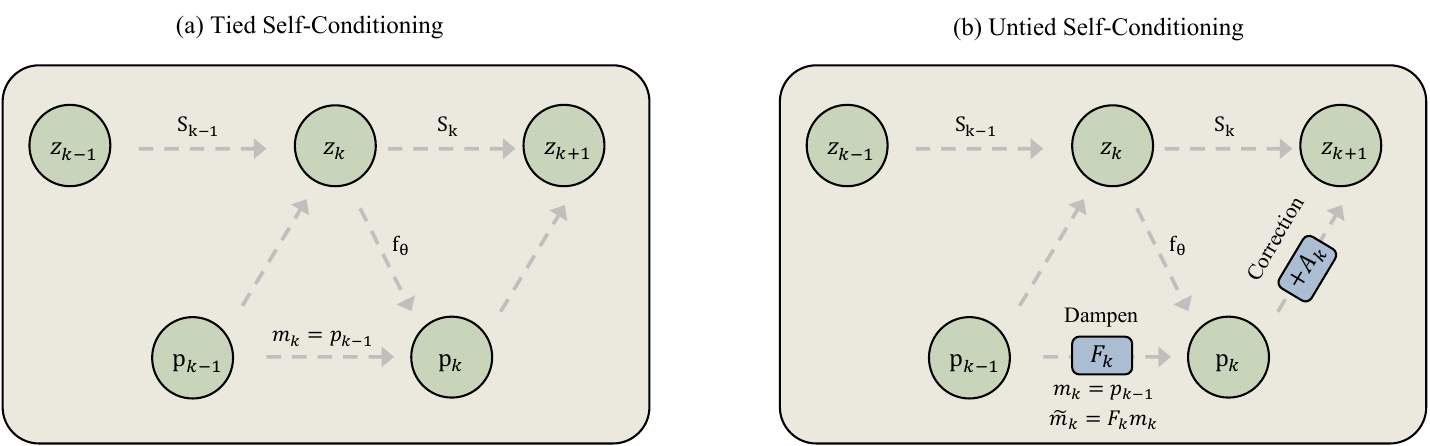}
\caption{\textbf{Train--inference mismatch and its correction.}
(a) Standard self-conditioning: the solver folds $p_{k-1}$ into the
latent $z_k$, so $z_k$ already encodes $p_{k-1}$.  Then $m_k=p_{k-1}$
re-introduces the same information as an explicit input.  During training,
no solver step intervenes, so this redundancy does not arise.
(b) \method{} corrects both consequences: an overlap transform $F_k$
dampens self-conditioning components the latent already represents, while
a solver correction $A_k$ shifts the prediction toward the step average.
The raw prediction is stored for the next self-conditioning input.}
\label{fig:transport_toy}
\end{figure}

In this work, we show that the solver-introduced coupling degrades few-step
generation through two specific mechanisms, and derive a correction for each
from the model's own structure.  First, the solver folds the previous
prediction into the latent, so the explicit self-conditioning input
re-introduces information the latent already carries; we show from the
model's frozen projection weights that this redundancy concentrates in
directions shared by both input pathways.  A controlled experiment confirms
the causal role of coupling: varying only the dependence between successive
steps while keeping each step's marginal distribution fixed, the benefit of
the self-conditioning input shrinks monotonically, and at strong coupling
full self-conditioning becomes harmful
(Section~\ref{sec:coupling_experiment}).  Second, the single-point prediction
at the start of each step departs from the average prediction needed for the
exact solver update; we show that this gap is first order in step width and
can be approximated from prediction history.  Both effects compound in the
few-step regime, explaining the sharp quality loss.

We introduce \method{}, a training-free sampler that corrects both mechanisms
with one model evaluation per step.  For the self-conditioning input, an
eigendecomposition of the frozen projection weights identifies the directions
along which the input is redundant with the latent; a transform dampens those
directions so the explicit input carries complementary information.  For the
solver, prediction history approximates the step-average prediction, with
scale set by offline trajectory statistics.  The corrected prediction enters
only the solver; the raw prediction is stored for self-conditioning.

Across ELF and LangFlow, \method{} improves generation quality in every
tested configuration.  At eight model evaluations it reduces LangFlow
GenPPL on OpenWebText from $531$ to~$62$ ($8.6\times$)
and on ELF-B from $71$ to~$43$.  At 32 evaluations on
ELF-B it achieves a GenPPL of~$21$, approaching the data reference,
with no retraining and $95\%$ throughput retained.

Our contributions are:
\begin{itemize}[leftmargin=1.4em,itemsep=0.25em,topsep=0.3em]
    \item We show that self-conditioning can become harmful at strong
    coupling, even when each step's prediction quality is unchanged.  The
    cause is a train--inference mismatch: the solver couples the latent and
    self-conditioning pathways in a way that is absent during training.
    A controlled experiment confirms the causal role by varying only the
    coupling while preserving marginals.
    \item We derive the geometric structure of the resulting redundancy from
    the model's frozen projection weights, and the solver's approximation
    gap from its integration structure; each derivation yields a specific
    correction objective.
    \item \method{} implements both corrections in a sampler that requires no
    retraining and uses one model evaluation per step.  It improves generation
    quality at every tested NFE, with the largest gains in the few-step
    regime.
\end{itemize}

Section~\ref{sec:preliminaries} reviews the sampling recurrence.
Section~\ref{sec:problem} diagnoses the mismatch and derives a correction
objective for each consequence.
Section~\ref{sec:method} assembles the complete sampler.
Section~\ref{sec:experiments} presents controlled experiments and generation
results.

\section{Preliminaries}
\label{sec:preliminaries}

A language flow embeds discrete tokens into a continuous space and
learns a velocity field $v_\theta(z,t)$ that transports noise to clean
data.  Sampling integrates the induced ODE from a noise sample~$z_0$;
in practice, a numerical solver discretizes the integral over a time
grid $0{=}t_0{<}\cdots{<}t_K{=}1$.  At each step~$k$, the model
produces a clean prediction~$p_k$ from which the solver computes the
latent update.  We report the total number of frozen-model evaluations as
NFE.

Previous-prediction self-conditioning~\citep{analog_bits} supplies the
model with its own clean prediction from the preceding step as an
additional input.  The sampling recurrence is
\begin{equation}
    p_k=f_\theta(z_k,m_k,t_k),\qquad
    z_{k+1}=S_k(z_k,p_k),\qquad
    m_{k+1}=p_k,
    \label{eq:vanilla_recurrence}
\end{equation}
where $m_k$ is the \emph{self-conditioning input}: the previous clean
prediction carried forward from step~$k{-}1$.  During training, the model
receives $m=\sg[f_\theta(z,0,t)]$ computed from the current noisy state
for a random half of examples, and zero
otherwise~\citep{analog_bits,sed,tess,plaid,seqdiffuseq,fmseq,elf}.
During sampling, $m_k{=}p_{k-1}$ while $z_k$ has already been advanced
with~$p_{k-1}$.  The intervening solver step is the source of cross-step
correlation analyzed in Section~\ref{sec:coupling_identifiability}.

\section{Solver Coupling Creates a Train--Inference Mismatch}
\label{sec:problem}
\label{sec:analysis}

During training, the model receives the current noisy state $z$ and a
self-conditioning input $m$ computed from that same state through a
detached forward pass.  No solver step intervenes between $z$ and $m$,
so no earlier prediction has been folded into $z$ by the solver.
During sampling, the situation is different.
The solver updates $z_{k+1}$ as a weighted combination of $z_k$ and
$p_k$, folding the prediction directly into the latent.  Then the next
evaluation sees $m_{k+1}=p_k$ as an explicit input alongside that updated
latent.  The model therefore receives $p_k$ twice: once absorbed into
$z_{k+1}$ through the solver, once passed directly as $m_{k+1}$.
This coupling between the two input pathways is absent during training
and grows with step width.

The mismatch has two consequences.
First, the self-conditioning input partly repeats what the latent already
encodes; the redundancy concentrates in directions that both input
projections share and can be quantified from the model's frozen weights.
Second, the solver evaluates the model once at the start of each step,
but accurate integration requires the average prediction over the step;
the gap grows at low step counts.
Section~\ref{sec:coupling_identifiability} establishes the causal role of
the coupling.
Sections~\ref{sec:conditional_innovation}
and~\ref{sec:interval_prediction} derive a correction objective for each
consequence.

\subsection{Cross-step correlation reduces the value of self-conditioning}
\label{sec:coupling_identifiability}

To confirm that the coupling itself, not the quality of individual
predictions, drives the degradation, we first isolate its effect in a
scalar Gaussian model.  The useful coefficient on the self-conditioning
input depends on how much of its information the latent already carries.
Let
$Y\sim\mathcal N(0,\tau^2)$ be a clean target observed through two noisy
channels:
\begin{equation}
    Z=Y+\sigma_z\epsilon_z,\qquad
    M=Y+\sigma_m\epsilon_m,\qquad
    \operatorname{Corr}(\epsilon_z,\epsilon_m)=\rho,
    \label{eq:gaussian_coupling}
\end{equation}
where $\epsilon_z,\epsilon_m$ are zero-mean, unit-variance, jointly
Gaussian, and independent of~$Y$.
The correlation $\rho$ controls how much noise the two channels share,
analogous to the solver coupling $p_{k-1}$ into both~$z_k$ and~$m_k$.
Varying $\rho$ leaves both marginals $P(Y,Z)$ and $P(Y,M)$
unchanged.  Only the joint $P(Z,M)$ changes.

The coefficient on $M$ that minimizes prediction error decreases
monotonically with~$\rho$, crossing zero at a critical correlation
that depends on the noise levels
(Appendix~\ref{app:coupling_proof}).  Beyond this crossing, a fixed
positive coefficient on $M$ becomes harmful.  So changing only the coupling between inputs can change whether the
self-conditioning input helps or hurts prediction.

The sampling recurrence has this structure.  The prediction $p_{k-1}$
appears explicitly in $m_k$ and also affects $z_k$ through the solver.
Section~\ref{sec:coupling_experiment} confirms the predicted monotone
decrease on a frozen language model.

\subsection{The self-conditioning input should reduce its overlap with the latent}
\label{sec:conditional_innovation}

Given that the coupling creates redundancy between the two input
pathways, we now derive its geometric structure from the model's
frozen projection weights.  The redundancy concentrates in directions
that both input projections share; this structure determines which
components of the self-conditioning input to dampen.

In self-conditioned flow-matching language models, the latent and self-conditioning input enter
the network through a single learned projection that operates on their
concatenation.  This is equivalent to two linear maps whose outputs are
summed.  Let $P_z$ and $P_m$ project onto the row spaces of these
effective projection matrices.
On the self-conditioning subspace, define
\begin{equation}
    K=P_mP_zP_m.
    \label{eq:canonical_operator}
\end{equation}
The eigenvectors $q_i$ of $K$ are the principal directions shared by the
two row spaces, and the eigenvalues $\rho_i^2\in[0,1]$ are the squared
cosines of their principal angles.  Concretely, a large~$\rho_i^2$ means
that both the latent projection and the self-conditioning projection
respond strongly to the same input component: any signal the
self-conditioning input carries along that direction is already
represented in the latent.  Along such directions the explicit input is
redundant.

For any self-conditioning input~$m$, the weighted projection onto these
shared directions is
\begin{equation}
    \mathcal E(m)=\tfrac{1}{2}\|P_zP_mm\|^2
    =\tfrac{1}{2}\ip{m}{Km}.
    \label{eq:overlap_energy}
\end{equation}
This quantity is our overlap objective: smaller $\mathcal E$ means less
redundancy with the latent pathway.

The objective $\mathcal E$ suggests a natural family of transforms:
dampen $m$ along high-overlap directions and leave the rest unchanged.
The map $(I{-}aK)$ with $0<a\le1$ does exactly this.%
\phantomsection\label{prop:overlap_contraction}%
  It retains fraction $1{-}a\rho_i^2$ along direction~$q_i$, so
directions shared by both projections ($\rho_i$ large) are dampened
most, while directions unique to self-conditioning pass through.
Section~\ref{sec:self_conditioning_transform} instantiates this family.

\subsection{The solver should approximate the step-average prediction}
\label{sec:interval_prediction}

The second consequence of the mismatch concerns the solver itself.
The solver evaluates the model once at the start of each step
$[t_k,t_{k+1}]$, but the true velocity field changes throughout the
step.  The exact update requires a weighted average $\bar p_k$
of predictions over the step interval
(Appendix~\ref{app:interval_proof}); a single evaluation provides only
the initial point~$p_k$.
\phantomsection\label{lem:finite_interval}%

The gap $C_k^*{=}\bar p_k{-}p_k$ is first order in step width~$h_k$,
and its contribution to the latent update is second order in~$h_k$:
negligible with many steps but substantial at low NFE.
Appendix~\ref{app:reference_experiments} verifies these orders on
high-resolution trajectories.

Computing $\bar p_k$ exactly requires multiple evaluations per step.
A weaker condition suffices:
if a nonzero direction $B_k$ is applied with scale $\eta>0$, the
change in squared correction error is
\begin{equation}
\|C_k^*-\eta B_k\|^2-\|C_k^*\|^2
  =\eta^2\|B_k\|^2-2\eta\ip{C_k^*}{B_k}.
\label{eq:positive_alignment_condition}
\end{equation}
The endpoint error is proportional to this quantity
(Appendix~\ref{app:interval_proof}), so any direction~$B_k$ positively
correlated with~$C_k^*$ improves the update for a range of
scales~$\eta$.

Computing $\bar p_k$ from past predictions avoids extra evaluations.
An exponential moving average tracks recent output:
\begin{equation}
    E_{k+1}=(1-\alpha)E_k+\alpha p_k,\qquad E_0=0.
    \label{eq:biased_history}
\end{equation}
The residual $H_k=p_k-E_k$ estimates recent change in prediction
and serves as the primary correction direction.
Section~\ref{sec:solver_correction} refines it with a second direction
from solver-weighted history, and normalizes the result using offline
trajectory statistics.
Appendix~\ref{app:reference_experiments} evaluates the resulting
direction against high-resolution step averages.

\section{\method{}}
\label{sec:method}

Section~\ref{sec:problem} diagnosed a train--inference mismatch with two
consequences: the self-conditioning input carries redundancy that
concentrates in specific directions of the projection geometry
(Section~\ref{sec:conditional_innovation}), and the solver's single-point
evaluation departs from the step-average prediction
(Section~\ref{sec:interval_prediction}).  We now implement a correction for
each within a single sampling loop.

A transform derived from the overlap structure dampens the
self-conditioning input along redundant directions, and a history-based
correction shifts the solver toward its step average.  Both share one model
evaluation per step; offline trajectory statistics set the correction's
mean and scale.

\subsection{Self-conditioning transform}
\label{sec:self_conditioning_transform}

The goal is to reduce $\mathcal E(m)$ from
Section~\ref{sec:conditional_innovation} at each step.  Principal
directions $Q$ and squared cosines $\rho_i^2$ are computed once from
frozen weights; the transform retains fraction $\ell_{k,i}$ along
direction~$q_i$:
\begin{equation}
\begin{aligned}
    F_k&=I-Q\!\left(I-\diag(\ell_{k,1},\ldots,\ell_{k,r})\right)Q^\top,\\
    \widetilde m_k&=F_km_k.
\end{aligned}
    \label{eq:released_self_conditioning}
\end{equation}
Directions with larger $\rho_i^2$ receive stronger dampening, following
the $(I{-}aK)$ form that reduces the overlap $\mathcal E$ at every step
(Section~\ref{sec:conditional_innovation}).  The base exponent $\lambda_R^{(0)}$ is derived from the model's
response geometry (Appendix~\ref{app:precomputation_details});
$\eta_R\ge0$ scales it, giving effective exponent
$\lambda_R=\lambda_R^{(0)}\eta_R$.  The dampening strength adapts across
steps: it is strongest when the solver has folded a large fraction of
the prior prediction into the latent, and weakens when the raw and
corrected running averages disagree (Section~\ref{sec:state_update}).
At the first step no prior prediction exists, so the input passes
unchanged.
Appendix~\ref{app:recurrent_transform} gives the full retention schedule.

\subsection{Solver-step correction}
\label{sec:solver_correction}

\paragraph{Solver-weighted history.}
Each solver step $S_k(z_k,q_k)$ depends linearly on the prediction~$q_k$
it receives (before correction $q_k{=}p_k$; with correction
$q_k{=}p_k{+}A_k$, see Section~\ref{sec:state_update}).  We track the
running weighted average of predictions that entered the solver:
\begin{equation}
    O_{k+1}=(1-\kappa_k)\,O_k+\kappa_k\,q_k,
    \qquad O_0=0,
    \label{eq:solver_weighted_state}
\end{equation}
where $\kappa_k\in[0,1]$ is the solver's update weight
(Appendix~\ref{app:solver_coordinates}).

\paragraph{Correction direction.}
We construct a direction that correlates with $C_k^*$ from two signals
available in the current trajectory.  The first is the EMA
residual $H_k{=}p_k{-}E_k$ from
Section~\ref{sec:interval_prediction}, which captures recent prediction
change.  The second compares the current prediction with the running
average~$O_k$:
\begin{equation}
    G_k=\begin{cases}\|H_k\|\,\dfrac{p_k-O_k}{\|p_k-O_k\|}&\text{if }\|H_k\|,\|p_k-O_k\|>\epsilon,\\[4pt]0&\text{otherwise}\end{cases}.
    \label{eq:direction_estimates}
\end{equation}
The combined correction direction is
\begin{equation}
    U_k=H_k+\eta_O\,\Delta u_k\,(H_k-G_k),
    \label{eq:combined_direction}
\end{equation}
where $\Delta u_k\in[0,1]$ is a normalized step-width feature
(Appendix~\ref{app:history_details}) and $\eta_O\ge0$ weights the
disagreement between the two estimates.
The two estimates reflect different prediction streams: $E_k$
accumulates the raw prediction~$p_k$ that enters self-conditioning,
while $O_k$ accumulates the corrected prediction $p_k{+}A_k$ that
enters the solver.  Their per-token cosine distance~$\bar\delta_k$
quantifies cross-pathway agreement and also gates the self-conditioning
dampening strength (Section~\ref{sec:state_update}).

\paragraph{Normalizing with offline statistics.}
The direction~$U_k$ comes from the current trajectory, but its scale is
arbitrary.  We normalize it to match statistics precomputed from stored
trajectories: a per-step token mean~$\mu_k$ and centered
RMS~$\sigma_k$, estimated on disjoint trajectory banks
(Appendix~\ref{app:empirical_carrier_statistics}).
Let $C_{\rm tok}$ center a tensor by subtracting its per-position mean.
The normalized correction is
\begin{equation}
\begin{aligned}
    \Pi_k(U_k)&=\mu_k+\sigma_k
        \frac{C_{\rm tok}U_k}{\RMS(C_{\rm tok}U_k)},\\
    A_k&=\eta_T\Pi_k(U_k).
\end{aligned}
    \label{eq:released_solver_correction}
\end{equation}
This replaces the scale and mean of~$U_k$ while preserving its centered
direction (Appendix~\ref{app:correction_details}).
Appendix~\ref{app:reference_experiments} evaluates the resulting
direction against high-resolution step averages.

\subsection{State update}
\label{sec:state_update}

One evaluation produces $p_k=f_\theta(z_k,\widetilde m_k,t_k)$; each
correction applies to its own pathway:
\begin{equation}
\boxed{
\begin{aligned}
    z_{k+1}&=S_k(z_k,p_k+A_k),
    &m_{k+1}&=p_k,\\
    O_{k+1}&=(1-\kappa_k)O_k+\kappa_k(p_k+A_k),
    &E_{k+1}&=(1-\alpha)E_k+\alpha p_k.
\end{aligned}}
    \label{eq:untied_state_update}
\end{equation}
The raw $p_k$ updates self-conditioning and the EMA; the corrected
$p_k+A_k$ advances the solver and its state.  During training the model
receives raw predictions as self-conditioning input; routing the
correction there would create a second mismatch
(Section~\ref{sec:assignment_experiments}).

The untied routing produces two running averages, $E_k$ from the raw
pathway and $O_k$ from the corrected pathway, whose agreement connects
both corrections.  The self-conditioning dampening adapts via
\begin{equation}
    a_k=\kappa_{k-1}(1-\bar\delta_{k-1}),
    \label{eq:dampening_gate}
\end{equation}
where $\bar\delta_{k-1}$ is the cosine distance between $H_{k-1}$
and~$G_{k-1}$.  When the estimates agree, dampening is strong; when
they disagree, it relaxes ($a_0{=}0$).
Algorithm~\ref{alg:method} summarizes the loop.

\begin{algorithm}[t]
\caption{\method{} sampling.  The corrected prediction updates the solver
states $(z,O)$; the raw prediction updates the self-conditioning states $(m,E)$.}
\label{alg:method}
\small
\textbf{Input:} frozen model $f_\theta$, base solver $S_k$, time grid
$\{t_k\}_{k=0}^{K}$

\textbf{Precomputed:} overlap basis $Q$, cosines $\{\rho_i^2\}$,
base transform exponent $\lambda_R^{(0)}$, cosine distance $\bar\delta_k$,
per-step correction mean $\mu_k$ and centered RMS $\sigma_k$

\textbf{Hyperparameters:} $\eta_R,\eta_T,\eta_O$ (nonnegative strength
coefficients); $\alpha\in(0,1]$ (EMA weight on the current prediction)

\textbf{Initialize:} $m_0,E_0,O_0\gets0$;
$\kappa_{-1}\gets0$, $\bar\delta_{-1}\gets0$

\textbf{for} $k=0,\ldots,K{-}1$ \textbf{do}
\begin{enumerate}[leftmargin=1.55em,label=\arabic*.,itemsep=0pt,topsep=1pt,parsep=0pt]
    \item $\widetilde m_k\gets F_km_k$
          \hfill\textit{self-conditioning transform}
    \item $p_k\gets f_\theta(z_k,\widetilde m_k,t_k)$
          \hfill\textit{one model evaluation}
    \item $H_k\gets p_k-E_k$; $G_k\gets
          \|H_k\|(p_k-O_k)/\|p_k-O_k\|$; construct $U_k$ by
          \eqref{eq:combined_direction}
    \item Compute $A_k$ by \eqref{eq:released_solver_correction}
          \hfill\textit{solver correction}
    \item $z_{k+1}\gets S_k(z_k,p_k+A_k)$; $O_{k+1}\gets
          (1-\kappa_k)O_k+\kappa_k(p_k+A_k)$
          \hfill\textit{corrected $\to$ solver}
    \item $m_{k+1}\gets p_k$; $E_{k+1}\gets
          (1-\alpha)E_k+\alpha p_k$
          \hfill\textit{raw $\to$ self-conditioning}
\end{enumerate}
\textbf{end for}; \textbf{return} $z_K$

\medskip\noindent
The loop uses exactly NFE evaluations: one per solver step for ELF;
NFE${-}1$ before the solver updates plus one final evaluation for LangFlow
(Appendix~\ref{app:solver_coordinates}).
\end{algorithm}

\section{Experiments}
\label{sec:experiments}

We first show that \method{} improves generation quality at every tested
NFE, then verify the diagnosed mismatch and each derived correction through
controlled experiments that isolate cross-step coupling, overlap-aligned
directions, and correction placement.

\subsection{Experimental setup}
\label{sec:setup}

\paragraph{Data and generation.}
ELF-B, ELF-M, and ELF-L are evaluated on OpenWebText with NFE
ranging from 8 to 64.  LangFlow is evaluated on OpenWebText and LM1B with
NFE from 8 to 256.  OpenWebText generations contain 1,024 tokens; LangFlow
LM1B generations contain 128.

\paragraph{Evaluation.}
We report generative perplexity (GenPPL) under GPT-2 Large and unigram
entropy~$H$.  GenPPL measures the quality of generated text; $H$ measures
diversity.  Section~\ref{sec:main_results} additionally reports pairwise
LLM-judge preference and per-sample external-evaluator likelihood as
complementary quality
signals.  All main results average six seeds, each with 1,024
generations; controlled studies use smaller sample counts noted in each
subsection.  Appendix~\ref{app:evaluation_protocol} gives the full
metrics, uncertainty units, and protocol.

\paragraph{Matched conditions.}
Every comparison keeps the pretrained model, base solver, time grid,
tokenizer, and decoding procedure fixed.  Matched configurations share the
initial latent, time grid, and random stream.  The overlap basis and offline trajectory
statistics are computed once from frozen model weights and shared across
all configurations.

\subsection{Generation quality improves at every tested NFE}
\label{sec:main_results}

\begin{figure}[t]
\centering
\includegraphics[width=0.45\linewidth]{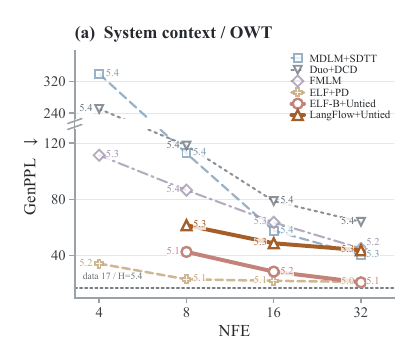}\hfill
\includegraphics[width=0.45\linewidth]{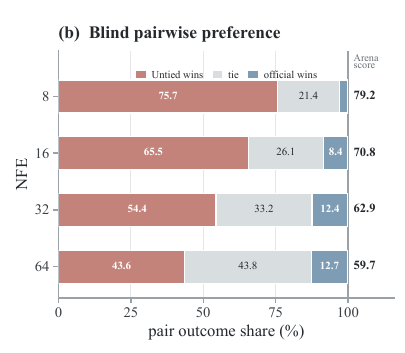}
\caption{\textbf{System context and pairwise LLM-judge preference.}
(a) System-level OpenWebText context; external curves taken from
\citet{elf}.
(b) Pair outcomes on LangFlow OpenWebText from 1,024 comparisons per
NFE\@.  Each bar partitions \methodshort{} wins, ties, and Official wins;
labels at the right endpoint give the Arena score.}
\label{fig:system_arena}
\end{figure}

\method{} improves generation quality relative to the official sampler in
every tested configuration
(Tables~\ref{tab:elf_main},~\ref{tab:langflow_main}). At 8~NFE, \method{} reduces LangFlow GenPPL on
OpenWebText from $531.2$ to $61.6$ ($8.6\times$) and on LM1B from $264.0$
to $73.0$ ($3.6\times$).  On ELF-B at 8~NFE it reaches $42.66$ versus the
official sampler's $70.63$.  At 32~NFE on ELF-B it reaches $21.0$
without additional training.
Figure~\ref{fig:system_arena}(a) places the results in system-level context.

\begin{figure}[t]
\centering
\includegraphics[width=\linewidth]{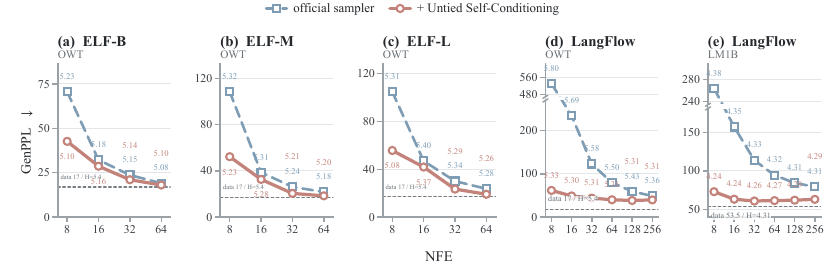}
\caption{\textbf{GenPPL versus NFE; every curve point annotates entropy.}
Panels (a)--(e) compare the official sampler (dashed) with \method{} (solid)
for each model--dataset pair.  Dotted lines mark data references taken from
\citet{elf} (OWT) and \citet{fixed_point_flows} (LM1B).}
\label{fig:main_results}
\end{figure}

Figure~\ref{fig:main_results} annotates unigram entropy at every matched
point: on LangFlow, \method{} moves entropy closer to the corpus
reference; on ELF, entropy is preserved.

At 8 and 16~NFE on LangFlow, every one of 6,144 paired samples has
lower GenPPL (Appendix~\ref{app:sample_quality}).  On ELF-B at 8~NFE,
varying each coefficient independently over a $[\times\!\frac{1}{2},\,
\times\!2]$ range always improves GenPPL
(Appendix~\ref{app:sensitivity}).  Across six workloads the sampler
retains $95.5\%$ geometric-mean throughput
(Appendix~\ref{app:throughput}).

\paragraph{Pairwise LLM-judge evaluation.}
Figure~\ref{fig:system_arena}(b) provides a direct quality comparison
independent of perplexity using an adapted Arena-Hard-Auto~v2
protocol~\citep{arena_hard_auto}.
Each NFE pairs 1,024 generations from the official sampler and
\method{} on the same latent; each pair is judged blind in both
presentation orders by DeepSeek~V4~Pro, yielding 8,192 judgments in total.
\method{} wins~$96.3\%$ of non-tie comparisons at 8~NFE (Arena
score~$79.2$, 95\% CI $[77.7,80.8]$) and retains a clear advantage at
64~NFE ($77.4\%$, Arena score~$59.7$, CI $[58.1,61.2]$).
Appendix~\ref{app:arena_protocol} gives the full protocol.

Appendix~\ref{app:additional_ablations} additionally compares \method{}
with training-free sampling methods under matched conditions;
\method{} outperforms all tested alternatives.
The following controlled experiments verify the diagnosed mismatch
and each derived correction independently.  The step-average scaling
experiment (Appendix~\ref{app:reference_experiments}) additionally verifies
that the correction $C_k^*$ is first order in step width and its latent
contribution is second order, matching
Section~\ref{sec:interval_prediction}.

\subsection{Does correlation across sampling steps change a fixed self-conditioning update?}
\label{sec:coupling_experiment}

\begin{figure}[t]
\centering
\begin{tikzpicture}
  \node[anchor=south west,inner sep=0] (temporal-base) at (0,0) {%
    \includegraphics[width=\linewidth]{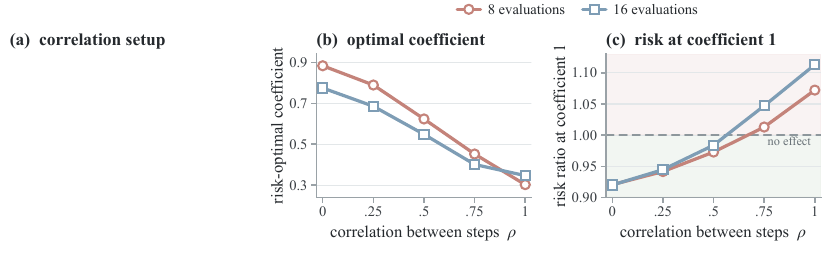}%
  };
  \begin{scope}[x={(temporal-base.south east)},y={(temporal-base.north west)}]
    \node[anchor=center,inner sep=0] at (0.16,0.50) {%
      \includegraphics[width=0.315\linewidth,trim=0 0 0 19pt,clip]{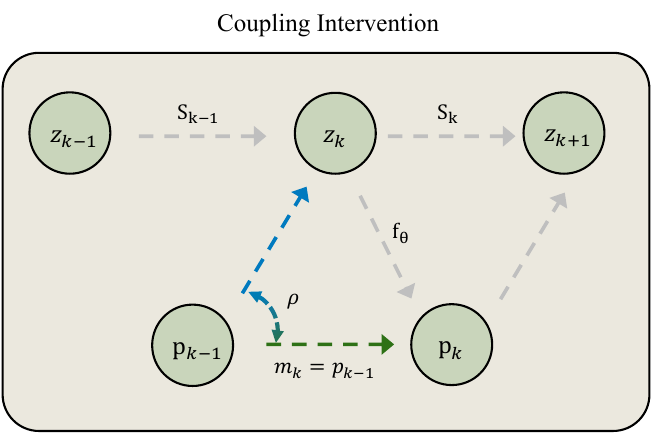}%
    };
  \end{scope}
\end{tikzpicture}
\caption{\textbf{Correlation across sampling steps changes the value of a fixed
self-conditioning update.}  (a) The experiment varies only the dependence
$\rho$ between the carried prediction
$m_k=p_{k-1}$ and the next latent $z_k$, while preserving both marginals.
The MSE-optimal interpolation coefficient along one frozen
self-conditioning-induced prediction direction decreases with $\rho$ (b);
consequently, a fixed coefficient of one crosses from helpful to harmful (c).}
\label{fig:temporal_correlation}
\end{figure}

This experiment directly tests the causal mechanism diagnosed in
Section~\ref{sec:coupling_identifiability}: does the coupling itself,
not the quality of individual predictions, change the value of
self-conditioning?  The experiment in
Figure~\ref{fig:temporal_correlation} preserves the marginal distribution at
every NFE and changes only the correlation between successive steps.  Five
correlations from $0$ to $1$ are tested at 8 and 16~NFE.

The MSE-optimal coefficient decreases monotonically from $0.88$ to $0.30$
at 8~NFE and from $0.78$ to $0.35$ at 16~NFE.  At high correlation, full
self-conditioning becomes harmful: the MSE ratio (coefficient one vs.\ zero)
rises from $0.92$ to $1.07$ and from $0.92$ to $1.11$, respectively.
The useful coefficient therefore depends on the joint distribution along the
sampling trajectory, confirming that the solver-introduced correlation
drives the effect.  Appendix~\ref{app:coupling_protocol} reports the full
protocol and controls.

\subsection{Which self-conditioning directions should be retained?}
\label{sec:spectrum_experiment}

Dampening high-overlap directions outperforms matched alternatives.
Table~\ref{tab:spectrum_pairing} (Appendix~\ref{app:spectrum_protocol})
holds eigenvalues fixed and varies only their assignment to input
directions, with the solver correction disabled.  The overlap-aligned,
reverse, and random transforms share the same eigenvalues and mean
strength; they differ only in which directions receive those eigenvalues.
The overlap-aligned assignment from
Section~\ref{sec:conditional_innovation} most strongly reduces
$\mathcal E(m)$ and outperforms official self-conditioning
by $0.15$ GPT-2 NLL as well as both matched controls.

\subsection{Where should the solver correction be applied?}
\label{sec:assignment_experiments}

\begin{table}[t]
\centering
\caption{Two $2\times2$ tests on ELF-B at 8~NFE with 1,024 matched
generations from seed~0 (GenPPL / unigram entropy):
(a) state assignment; (b) method components.}
\label{tab:placement}
\footnotesize
\begin{minipage}[t]{0.53\linewidth}
\centering
\textbf{(a) State assignment}\\[3pt]
\begin{tabular*}{\linewidth}{@{\extracolsep{\fill}}lcc@{}}
\toprule
& Solver: $p_k$ & Solver: $p_k+A_k$ \\
\midrule
Next SC: $p_k$     & 67.27 / 5.23  & \textbf{45.63 / 5.14} \\
Next SC: $p_k+A_k$ & 157.77 / 5.35 & 334.19 / 5.33 \\
\bottomrule
\end{tabular*}
\end{minipage}\hfill
\begin{minipage}[t]{0.44\linewidth}
\centering
\textbf{(b) Components}\\[3pt]
\begin{tabular*}{\linewidth}{@{\extracolsep{\fill}}lcc@{}}
\toprule
SC transform & Correction off & Correction on \\
\midrule
Off & 74.18 / 5.27 & 50.04 / 5.16 \\
On  & 67.27 / 5.23 & \textbf{45.63 / 5.14} \\
\bottomrule
\end{tabular*}
\end{minipage}
\end{table}

The solver is the correct destination for the history-based correction.
Table~\ref{tab:placement} applies the same correction either in the solver
update, in the next self-conditioning input, or in both places, while
fixing the self-conditioning transform.  Applying it in the solver reaches
$45.63$ GenPPL compared with $67.27$ without the correction under the same
self-conditioning transform.  Adding the correction only to the
self-conditioning input raises GenPPL to $157.77$, worse than no
correction at all.

This ordering holds in five additional settings.  Panel~(b) isolates the
two components: the solver correction provides the larger individual gain,
and the complete sampler outperforms either component alone.
Appendix~\ref{app:additional_ablations} reports the additional settings and
direction controls.


\section{Related Work}
\label{sec:related_work}

\paragraph{Self-conditioned language flows.}
Continuous diffusion and flow models for text embed discrete tokens into a
continuous space and refine all positions in
parallel~\citep{diffusion_lm,sed,plaid,ssd_lm,tess,tess2,diffuseq,seqdiffuseq,fmseq,replaid,flowseq,langflow,elf}.
Previous-prediction self-conditioning~\citep{analog_bits} supplies the model
with its own clean prediction from the preceding step; variants appear
across embedding, simplex, and contextual
formulations~\citep{sed,tess,plaid,seqdiffuseq,fmseq,elf}.  We study models
in which a frozen network receives the current latent and a previous clean
prediction through separate linear input projections.  ELF and LangFlow are
two instances of this interface.

\paragraph{Training-time modifications to self-conditioning.}
Several studies diagnose train--inference mismatch in the self-conditioning
recurrence and address it through modified
training~\citep{tencdm,trec,fastdiss}.  Other methods expose the model to its
own intermediate states during training~\citep{step_unrolled}, specialize
through post-training~\citep{scmdm}, or distill recurrent self-conditioning
iterations~\citep{fixed_point_flows}.  All require modifying model weights.
Our method keeps the model frozen and constructs a transform from the geometry
of the input projections.

\paragraph{Inference-time changes to self-conditioning.}
Analog Bits studies an exponential moving average of previous predictions and a
self-guidance rule that uses two model evaluations~\citep{analog_bits}.  ACE
estimates a repetition-associated direction from generated trajectories and
subtracts it from the self-conditioning input~\citep{ace}.  These methods
modify the self-conditioning input independently of the solver update.

\paragraph{Solvers and multi-step methods.}
Advanced solvers reuse past model evaluations to improve numerical integration
of the probability-flow
ODE~\citep{deis,dpm_solver_pp,unipc}.  Separately, several methods aggregate
past clean predictions to stabilize or refine the current prediction supplied
to the solver~\citep{diffusion_momentum,masf,difa,higs,self_guidance}.  In
self-conditioned models, the same prediction enters both the solver and the
next network evaluation.  These methods improve the numerical update but do
not account for the self-conditioning pathway.

\paragraph{Offline model statistics for solver design.}
DPM-Solver-v3 precomputes evaluator-free empirical model statistics to
instantiate a model-dependent solver
parameterization~\citep{dpm_solver_v3}.  Offline trajectory statistics have
proved useful for training-free solver design more broadly.  These methods
target numerical integration without modifying the self-conditioning pathway.

In summary, prior methods modify individual components of the sampling loop
in isolation.  This work identifies that the solver introduces coupling
between the two pathways that is absent during training, and derives both
corrections from the model's projection geometry and integration structure.

\section{Conclusion}
\label{sec:conclusion}

We diagnosed a train--inference mismatch in self-conditioned flow sampling:
the solver introduces coupling between the latent and self-conditioning
pathways that is absent during training.  A controlled experiment confirmed
the causal role of this coupling: when it is strong, a fixed self-conditioning
coefficient crosses from helpful to harmful.  From the model's frozen
projection weights we derived the geometric structure of the resulting
redundancy, and from the solver's integration structure we derived the
step-average approximation gap.  \method{} corrects both in a training-free
sampler with one model evaluation per step.  Across ELF and LangFlow, it
improves generation quality at every tested NFE; with eight evaluations, it
reduces LangFlow GenPPL on OpenWebText from $531$ to~$62$, with gains
confirmed by pairwise LLM-judge preference and per-sample external-evaluator
likelihood.

\bibliography{references}

@inproceedings{
analog_bits,
title={Analog Bits: Generating Discrete Data using Diffusion Models with Self-Conditioning},
author={Ting Chen and Ruixiang Zhang and Geoffrey Hinton},
booktitle={The Eleventh International Conference on Learning Representations },
year={2023},
url={https://openreview.net/forum?id=3itjR9QxFw}
}

@InProceedings{arena_hard_auto,
  title = 	 {From Crowdsourced Data to High-quality Benchmarks: Arena-Hard and Benchbuilder Pipeline},
  author =       {Li, Tianle and Chiang, Wei-Lin and Frick, Evan and Dunlap, Lisa and Wu, Tianhao and Zhu, Banghua and Gonzalez, Joseph E. and Stoica, Ion},
  booktitle = 	 {Proceedings of the 42nd International Conference on Machine Learning},
  pages = 	 {34209--34231},
  year = 	 {2025},
  editor = 	 {Singh, Aarti and Fazel, Maryam and Hsu, Daniel and Lacoste-Julien, Simon and Berkenkamp, Felix and Maharaj, Tegan and Wagstaff, Kiri and Zhu, Jerry},
  volume = 	 {267},
  series = 	 {Proceedings of Machine Learning Research},
  month = 	 {13--19 Jul},
  publisher =    {PMLR},
  url = 	 {https://proceedings.mlr.press/v267/li25h.html}
}

@inproceedings{
deis,
title={Fast Sampling of Diffusion Models with Exponential Integrator},
author={Qinsheng Zhang and Yongxin Chen},
booktitle={The Eleventh International Conference on Learning Representations },
year={2023},
url={https://openreview.net/forum?id=Loek7hfb46P}
}

@inproceedings{
difa,
title={Di{FA}: Inference-Time Forward-Process Alignment for Diffusion Models},
author={Shigui Li and Delu zeng},
booktitle={Forty-third International Conference on Machine Learning},
year={2026},
url={https://openreview.net/forum?id=YGyWLM0OEW}
}

@inproceedings{
diffuseq,
title={DiffuSeq: Sequence to Sequence Text Generation with Diffusion Models},
author={Shansan Gong and Mukai Li and Jiangtao Feng and Zhiyong Wu and Lingpeng Kong},
booktitle={The Eleventh International Conference on Learning Representations },
year={2023},
url={https://openreview.net/forum?id=jQj-_rLVXsj}
}

@inproceedings{diffusion_lm,
 author = {Li, Xiang and Thickstun, John and Gulrajani, Ishaan and Liang, Percy S and Hashimoto, Tatsunori B},
 booktitle = {Advances in Neural Information Processing Systems},
 doi = {10.52202/068431-0313},
 editor = {S. Koyejo and S. Mohamed and A. Agarwal and D. Belgrave and K. Cho and A. Oh},
 pages = {4328--4343},
 publisher = {Curran Associates, Inc.},
 title = {Diffusion-LM Improves Controllable Text Generation},
 url = {https://proceedings.neurips.cc/paper_files/paper/2022/file/1be5bc25d50895ee656b8c2d9eb89d6a-Paper-Conference.pdf},
 volume = {35},
 year = {2022}
}

@inproceedings{
diffusion_momentum,
title={Diffusion Sampling with Momentum for Mitigating Divergence Artifacts},
author={Suttisak Wizadwongsa and Worameth Chinchuthakun and Pramook Khungurn and Amit Raj and Supasorn Suwajanakorn},
booktitle={The Twelfth International Conference on Learning Representations},
year={2024},
url={https://openreview.net/forum?id=HXc5aXeoc8}
}

@article{dpm_solver_pp,
   title={DPM-Solver++: Fast Solver for Guided Sampling of Diffusion Probabilistic Models},
   volume={22},
   ISSN={2731-5398},
   url={http://dx.doi.org/10.1007/s11633-025-1562-4},
   DOI={10.1007/s11633-025-1562-4},
   number={4},
   journal={Machine Intelligence Research},
   publisher={Springer Science and Business Media LLC},
   author={Lu, Cheng and Zhou, Yuhao and Bao, Fan and Chen, Jianfei and Li, Chongxuan and Zhu, Jun},
   year={2025},
   month=jun, pages={730--751}
}

@inproceedings{dpm_solver_v3,
 author = {Zheng, Kaiwen and Lu, Cheng and Chen, Jianfei and Zhu, Jun},
 booktitle = {Advances in Neural Information Processing Systems},
 doi = {10.52202/075280-2423},
 editor = {A. Oh and T. Naumann and A. Globerson and K. Saenko and M. Hardt and S. Levine},
 pages = {55502--55542},
 publisher = {Curran Associates, Inc.},
 title = {DPM-Solver-v3: Improved Diffusion ODE Solver with Empirical Model Statistics},
 url = {https://proceedings.neurips.cc/paper_files/paper/2023/file/ada8de994b46571bdcd7eeff2d3f9cff-Paper-Conference.pdf},
 volume = {36},
 year = {2023}
}

@inproceedings{fastdiss,
    title = "{F}ast{D}i{SS}: Few-step Match Many-step Diffusion Language Model on Sequence-to-Sequence Generation",
    author = "Cong, Dat Nguyen  and
      Kieu, Tung  and
      Thanh-Tung, Hoang",
    editor = "Liakata, Maria  and
      Moreira, Viviane P.  and
      Zhang, Jiajun  and
      Jurgens, David",
    booktitle = "Findings of the {A}ssociation for {C}omputational {L}inguistics: {ACL} 2026",
    month = jul,
    year = "2026",
    address = "San Diego, California, United States",
    publisher = "Association for Computational Linguistics",
    url = "https://aclanthology.org/2026.findings-acl.870/",
    doi = "10.18653/v1/2026.findings-acl.870",
    pages = "17572--17592",
    ISBN = "979-8-89176-395-1"
}

@inproceedings{
flow_matching,
title={Flow Matching for Generative Modeling},
author={Yaron Lipman and Ricky T. Q. Chen and Heli Ben-Hamu and Maximilian Nickel and Matthew Le},
booktitle={The Eleventh International Conference on Learning Representations },
year={2023},
url={https://openreview.net/forum?id=PqvMRDCJT9t}
}

@inproceedings{flowseq,
    title = "Flow Matching for Conditional Text Generation in a Few Sampling Steps",
    author = {Hu, Vincent  and
      Wu, Di  and
      Asano, Yuki  and
      Mettes, Pascal  and
      Fernando, Basura  and
      Ommer, Bj{\"o}rn  and
      Snoek, Cees},
    editor = "Graham, Yvette  and
      Purver, Matthew",
    booktitle = "Proceedings of the 18th Conference of the European Chapter of the Association for Computational Linguistics (Volume 2: Short Papers)",
    month = mar,
    year = "2024",
    address = "St. Julian{'}s, Malta",
    publisher = "Association for Computational Linguistics",
    url = "https://aclanthology.org/2024.eacl-short.33/",
    doi = "10.18653/v1/2024.eacl-short.33",
    pages = "380--392"
}

@inproceedings{fmseq,
    title = "Enable Fast Sampling for {S}eq2{S}eq Text Diffusion",
    author = "Liu, Pan  and
      Tian, Xiaohua  and
      Lin, Zhouhan",
    editor = "Al-Onaizan, Yaser  and
      Bansal, Mohit  and
      Chen, Yun-Nung",
    booktitle = "Findings of the Association for Computational Linguistics: EMNLP 2024",
    month = nov,
    year = "2024",
    address = "Miami, Florida, USA",
    publisher = "Association for Computational Linguistics",
    url = "https://aclanthology.org/2024.findings-emnlp.497/",
    doi = "10.18653/v1/2024.findings-emnlp.497",
    pages = "8495--8505"
}

@inproceedings{
higs,
title={Hi{GS}: History-Guided Sampling for Plug-and-Play Enhancement of Diffusion Models},
author={Seyedmorteza Sadat and Farnood Salehi and Romann M. Weber},
booktitle={The Fourteenth International Conference on Learning Representations},
year={2026},
url={https://openreview.net/forum?id=cyQUZDMpg3}
}

@InProceedings{masf,
    author    = {Qian, Yurui and Cai, Qi and Pan, Yingwei and Li, Yehao and Yao, Ting and Sun, Qibin and Mei, Tao},
    title     = {Boosting Diffusion Models with Moving Average Sampling in Frequency Domain},
    booktitle = {Proceedings of the IEEE/CVF Conference on Computer Vision and Pattern Recognition (CVPR)},
    month     = {June},
    year      = {2024},
    pages     = {8911-8920},
    url       = {https://openaccess.thecvf.com/content/CVPR2024/html/Qian_Boosting_Diffusion_Models_with_Moving_Average_Sampling_in_Frequency_Domain_CVPR_2024_paper.html}
}

@inproceedings{plaid,
 author = {Gulrajani, Ishaan and Hashimoto, Tatsunori B},
 booktitle = {Advances in Neural Information Processing Systems},
 doi = {10.52202/075280-0730},
 editor = {A. Oh and T. Naumann and A. Globerson and K. Saenko and M. Hardt and S. Levine},
 pages = {16693--16715},
 publisher = {Curran Associates, Inc.},
 title = {Likelihood-Based Diffusion Language Models},
 url = {https://proceedings.neurips.cc/paper_files/paper/2023/file/35b5c175e139bff5f22a5361270fce87-Paper-Conference.pdf},
 volume = {36},
 year = {2023}
}

@ARTICLE{self_guidance,
  author={Li, Tiancheng and Luo, Weijian and Chen, Zhiyang and Ma, Liyuan and Qi, Guo-Jun},
  journal={IEEE Transactions on Pattern Analysis and Machine Intelligence}, 
  title={Self-Guidance: Boosting Flow and Diffusion Generation on Their Own}, 
  year={2026},
  volume={48},
  number={1},
  pages={781-791},
  doi={10.1109/TPAMI.2025.3611831}}

@inproceedings{seqdiffuseq,
    title = "Text Diffusion Model with Encoder-Decoder Transformers for Sequence-to-Sequence Generation",
    author = "Yuan, Hongyi  and
      Yuan, Zheng  and
      Tan, Chuanqi  and
      Huang, Fei  and
      Huang, Songfang",
    editor = "Duh, Kevin  and
      Gomez, Helena  and
      Bethard, Steven",
    booktitle = "Proceedings of the 2024 Conference of the North American Chapter of the Association for Computational Linguistics: Human Language Technologies (Volume 1: Long Papers)",
    month = jun,
    year = "2024",
    address = "Mexico City, Mexico",
    publisher = "Association for Computational Linguistics",
    url = "https://aclanthology.org/2024.naacl-long.2/",
    doi = "10.18653/v1/2024.naacl-long.2",
    pages = "22--39"
}

@inproceedings{ssd_lm,
    title = "{SSD}-{LM}: Semi-autoregressive Simplex-based Diffusion Language Model for Text Generation and Modular Control",
    author = "Han, Xiaochuang  and
      Kumar, Sachin  and
      Tsvetkov, Yulia",
    editor = "Rogers, Anna  and
      Boyd-Graber, Jordan  and
      Okazaki, Naoaki",
    booktitle = "Proceedings of the 61st Annual Meeting of the Association for Computational Linguistics (Volume 1: Long Papers)",
    month = jul,
    year = "2023",
    address = "Toronto, Canada",
    publisher = "Association for Computational Linguistics",
    url = "https://aclanthology.org/2023.acl-long.647/",
    doi = "10.18653/v1/2023.acl-long.647",
    pages = "11575--11596"
}

@inproceedings{
step_unrolled,
title={Step-unrolled Denoising Autoencoders for Text Generation},
author={Nikolay Savinov and Junyoung Chung and Mikolaj Binkowski and Erich Elsen and Aaron van den Oord},
booktitle={International Conference on Learning Representations},
year={2022},
url={https://openreview.net/forum?id=T0GpzBQ1Fg6}
}

@misc{tencdm,
      title={TEncDM: Understanding the Properties of the Diffusion Model in the Space of Language Model Encodings}, 
      author={Alexander Shabalin and Viacheslav Meshchaninov and Egor Chimbulatov and Vladislav Lapikov and Roman Kim and Grigory Bartosh and Dmitry Molchanov and Sergey Markov and Dmitry Vetrov},
      year={2025},
      eprint={2402.19097},
      archivePrefix={arXiv},
      primaryClass={cs.CL},
      url={https://arxiv.org/abs/2402.19097}, 
}

@inproceedings{tess,
    title = "{TESS}: Text-to-Text Self-Conditioned Simplex Diffusion",
    author = "Karimi Mahabadi, Rabeeh  and
      Ivison, Hamish  and
      Tae, Jaesung  and
      Henderson, James  and
      Beltagy, Iz  and
      Peters, Matthew  and
      Cohan, Arman",
    editor = "Graham, Yvette  and
      Purver, Matthew",
    booktitle = "Proceedings of the 18th Conference of the European Chapter of the Association for Computational Linguistics (Volume 1: Long Papers)",
    month = mar,
    year = "2024",
    address = "St. Julian{'}s, Malta",
    publisher = "Association for Computational Linguistics",
    url = "https://aclanthology.org/2024.eacl-long.144/",
    doi = "10.18653/v1/2024.eacl-long.144",
    pages = "2347--2361"
}

@inproceedings{tess2,
    title = "{TESS} 2: A Large-Scale Generalist Diffusion Language Model",
    author = "Tae, Jaesung  and
      Ivison, Hamish  and
      Kumar, Sachin  and
      Cohan, Arman",
    editor = "Che, Wanxiang  and
      Nabende, Joyce  and
      Shutova, Ekaterina  and
      Pilehvar, Mohammad Taher",
    booktitle = "Proceedings of the 63rd Annual Meeting of the Association for Computational Linguistics (Volume 1: Long Papers)",
    month = jul,
    year = "2025",
    address = "Vienna, Austria",
    publisher = "Association for Computational Linguistics",
    url = "https://aclanthology.org/2025.acl-long.1029/",
    doi = "10.18653/v1/2025.acl-long.1029",
    pages = "21171--21188",
    ISBN = "979-8-89176-251-0"
}

@misc{trec,
      title={Text Diffusion with Reinforced Conditioning}, 
      author={Yuxuan Liu and Tianchi Yang and Shaohan Huang and Zihan Zhang and Haizhen Huang and Furu Wei and Weiwei Deng and Feng Sun and Qi Zhang},
      year={2024},
      eprint={2402.14843},
      archivePrefix={arXiv},
      primaryClass={cs.CL},
      url={https://arxiv.org/abs/2402.14843}, 
}

@inproceedings{unipc,
 author = {Zhao, Wenliang and Bai, Lujia and Rao, Yongming and Zhou, Jie and Lu, Jiwen},
 booktitle = {Advances in Neural Information Processing Systems},
 doi = {10.52202/075280-2170},
 editor = {A. Oh and T. Naumann and A. Globerson and K. Saenko and M. Hardt and S. Levine},
 pages = {49842--49869},
 publisher = {Curran Associates, Inc.},
 title = {UniPC: A Unified Predictor-Corrector Framework for Fast Sampling of Diffusion Models},
 url = {https://proceedings.neurips.cc/paper_files/paper/2023/file/9c2aa1e456ea543997f6927295196381-Paper-Conference.pdf},
 volume = {36},
 year = {2023}
}

@misc{ace,
      title={Low Perplexity is Repetition: A One-Dimensional Self-Conditioning Attractor in Continuous Diffusion LMs}, 
      author={Shuai Zhang and Zijie Chen and Hongliang He and Lun Du and Zhenzhong Lan},
      year={2026},
      eprint={2607.00588},
      archivePrefix={arXiv},
      primaryClass={cs.CL},
      url={https://arxiv.org/abs/2607.00588}, 
}

@misc{elf,
      title={ELF: Embedded Language Flows}, 
      author={Keya Hu and Linlu Qiu and Yiyang Lu and Hanhong Zhao and Tianhong Li and Yoon Kim and Jacob Andreas and Kaiming He},
      year={2026},
      eprint={2605.10938},
      archivePrefix={arXiv},
      primaryClass={cs.CL},
      url={https://arxiv.org/abs/2605.10938}, 
}

@misc{fixed_point_flows,
      title={Self-conditioned Flow Map Language Models via Fixed-point Flows}, 
      author={Jaehoon Yoo and Wonjung Kim and Floor Eijkelboom and Chanhyuk Lee and Nicholas M. Boffi and Seunghoon Hong and Jinwoo Kim},
      year={2026},
      eprint={2607.00714},
      archivePrefix={arXiv},
      primaryClass={cs.CL},
      url={https://arxiv.org/abs/2607.00714}, 
}

@misc{langflow,
      title={LangFlow: Continuous Diffusion Rivals Discrete in Language Modeling}, 
      author={Yuxin Chen and Chumeng Liang and Hangke Sui and Ruihan Guo and Chaoran Cheng and Jiaxuan You and Ge Liu},
      year={2026},
      eprint={2604.11748},
      archivePrefix={arXiv},
      primaryClass={cs.CL},
      url={https://arxiv.org/abs/2604.11748}, 
}

@misc{replaid,
      title={Continuous Diffusion Scales Competitively with Discrete Diffusion for Language}, 
      author={Zhihan Yang and Wei Guo and Shuibai Zhang and Subham Sekhar Sahoo and Yongxin Chen and Arash Vahdat and Morteza Mardani and John Thickstun},
      year={2026},
      eprint={2605.18530},
      archivePrefix={arXiv},
      primaryClass={cs.CL},
      url={https://arxiv.org/abs/2605.18530}, 
}

@misc{scmdm,
      title={Simple Self-Conditioning Adaptation for Masked Diffusion Models}, 
      author={Michael Cardei and Huu Binh Ta and Ferdinando Fioretto},
      year={2026},
      eprint={2604.26985},
      archivePrefix={arXiv},
      primaryClass={cs.LG},
      url={https://arxiv.org/abs/2604.26985}, 
}

@misc{sed,
      title={Self-conditioned Embedding Diffusion for Text Generation}, 
      author={Robin Strudel and Corentin Tallec and Florent Altché and Yilun Du and Yaroslav Ganin and Arthur Mensch and Will Grathwohl and Nikolay Savinov and Sander Dieleman and Laurent Sifre and Rémi Leblond},
      year={2022},
      eprint={2211.04236},
      archivePrefix={arXiv},
      primaryClass={cs.CL},
      url={https://arxiv.org/abs/2211.04236}, 
}
\bibliographystyle{preprint_style}

\appendix
\section{Derivations and Proofs}
\label{app:response_proofs}

\subsection{Gaussian conditioning calculation}
\label{app:coupling_proof}

Let $T=\tau^2$, $a=\sigma_z$, and $b=\sigma_m$.  Gaussian conditioning on
$(Z,M)$ gives the coefficient on~$M$ as
\begin{equation}
 b_\rho=\frac{T(a^2-\rho ab)}{\Delta_\rho},\qquad
 \Delta_\rho=(T+a^2)(T+b^2)-(T+\rho ab)^2>0.
 \label{eq:gaussian_conditioning_coefficient}
\end{equation}
Neither marginal $P_{Y,Z}$ nor $P_{Y,M}$ depends on~$\rho$.  For
$0<a<b$, direct differentiation gives
\begin{equation}
 \frac{\dd b_\rho}{\dd\rho}
 =-\frac{Tab\left[(b^2-a^2)(T+a^2)
                  +a^2b^2(\rho-a/b)^2\right]}{\Delta_\rho^2}<0.
 \label{eq:gaussian_coefficient_monotonicity}
\end{equation}
Thus $b_\rho$ vanishes at $\rho=a/b$ and changes sign there.  At
$\rho=1$, the shared-noise observations permit exact cancellation,
$Y=(bZ-aM)/(b-a)$.  The conditional fusion rule can therefore change sign
while both feature--target marginals remain fixed.

\subsection{Exact average of past predictions using the solver coefficients}
\label{app:solver_coordinates}

For ELF, distinguish the time $s_k$ of the model evaluation (after any
re-noising) from the target time $t_{k+1}$.  Write the latent at that evaluation
as
\begin{equation}
    z_k=(1-s_k)\nu_k+s_kO_k,
    \label{eq:elf_state_decomposition}
\end{equation}
where $\nu_k$ contains the complementary noise component and $O_k$ is the
normalized contribution of predictions previously supplied to the solver.
Suppose the solver uses $q_k$ to advance the latent from $s_k$ to $t_{k+1}$,
with
\begin{equation}
    \beta_k=\frac{t_{k+1}-s_k}{1-s_k}.
\end{equation}
Substitution into the affine solver update gives
\begin{equation}
    O_{k+1}=
    \frac{(1-\beta_k)s_kO_k+\beta_kq_k}{t_{k+1}}
    =(1-\kappa_k)O_k+\kappa_kq_k,
    \qquad
    \kappa_k=\frac{\beta_k}{t_{k+1}}.
    \label{eq:elf_solver_state}
\end{equation}
The identity
$(1-\beta_k)s_k+\beta_k=t_{k+1}$ makes the two coefficients sum to one.
The coefficient of $q_k$ in the full latent is $\beta_k$, whereas
$\kappa_k=\beta_k/t_{k+1}$ is its update share inside normalized~$O_{k+1}$.
With $O_0=0$ and $s_0=0$, the first normalized update has $\kappa_0=1$ and
$O_1=q_0$, even though the full-latent coefficient is only
$\beta_0=t_1$.  Re-noising changes $s_k$ but preserves this algebraic
separation.

For a LangFlow transition from negative log-SNR $\gamma_t$ to $\gamma_s$,
the clean-prediction component after dividing by the signal coefficient obeys
\begin{equation}
    O_s=e^{(\gamma_s-\gamma_t)/2}O_t
       +\left(1-e^{(\gamma_s-\gamma_t)/2}\right)q_t.
    \label{eq:langflow_solver_state}
\end{equation}
Thus \eqref{eq:solver_weighted_state} holds with
$\kappa=1-e^{(\gamma_s-\gamma_t)/2}$.  Additive stochastic terms belong to
the complementary component of the latent and do not enter $O_k$.

\subsection{Weighted prediction over a solver step and step-size orders}
\label{app:interval_proof}

Hold the self-conditioning input fixed between two model evaluations, and
let $y(u)$ denote the normalized latent coordinate driven by the clean
prediction.  For self-conditioned language flows whose solver updates the
latent affinely in the clean prediction, a change of time coordinate gives
\begin{equation}
    \frac{\dd y}{\dd u}=p(u)-y(u),
    \qquad
    p(u)=f_\theta(z(u),m_k,t(u)).
    \label{eq:conditional_flow}
\end{equation}
Here $u$ is the increasing solver time coordinate (ELF uses
$u=-\log(1-t)$ in its active region; LangFlow uses $u=-\gamma/2$) and
$h_k=u_{k+1}-u_k>0$ is the
width of step $k$ in this coordinate.
Multiplying \eqref{eq:conditional_flow} by $e^u$ gives
\begin{equation}
    \frac{\dd}{\dd u}\left(e^uy(u)\right)=e^up(u).
\end{equation}
Integration over $[u_k,u_k+h]$ yields
\begin{equation}
    y(u_k+h)=e^{-h}y(u_k)
      +\int_0^h e^{-(h-s)}p(u_k+s)\,\dd s.
\end{equation}
Defining the step-average prediction
\begin{equation}
    \bar p_k(h_k)
      =\frac{1}{1-e^{-h_k}}
        \int_0^{h_k}e^{-(h_k-s)}p(u_k+s)\,\dd s,
    \label{eq:interval_prediction}
\end{equation}
the exact endpoint is
\begin{equation}
    y(u_k+h_k)
      =e^{-h_k}y(u_k)
       +(1-e^{-h_k})\bar p_k(h_k).
    \label{eq:exact_interval_endpoint}
\end{equation}

If the prediction path is three times differentiable,
\begin{equation}
    p(u_k+s)=p_k+s\dot p_k+\frac{s^2}{2}\ddot p_k+O(s^3).
\end{equation}
Substitution into \eqref{eq:interval_prediction} gives
\begin{equation}
    C_k^*(h)
      =a_1(h)\dot p_k+\frac{a_2(h)}{2}\ddot p_k+O(h^3),
    \label{eq:second_order_interval_expansion}
\end{equation}
where
\begin{equation}
\begin{aligned}
    a_1(h)&=\frac{h-1+e^{-h}}{1-e^{-h}}
           =\frac h2+O(h^2),\\
    a_2(h)&=\frac{h^2-2h+2-2e^{-h}}{1-e^{-h}}
           =\frac{h^2}{3}+O(h^3).
\end{aligned}
\end{equation}
Since $1-e^{-h}=h+O(h^2)$, the true correction is first order and
its contribution to the endpoint is second order.
For any candidate $B_k$, substituting $p_k+\eta B_k$ for the interval
prediction in~\eqref{eq:exact_interval_endpoint} gives
\begin{equation}
 y_{k+1}^{(\eta B)}-y_{k+1}^{*}
 =(1-e^{-h_k})(\eta B_k-C_k^*).
\end{equation}
The squared endpoint error therefore differs
from~\eqref{eq:positive_alignment_condition} by the positive factor
$(1-e^{-h_k})^2$, preserving the sign of the improvement criterion.

The same weighted-average representation holds for a general scalar affine
flow
\begin{equation}
    \frac{\dd y}{\dd u}=a(u)y(u)+b(u)p(u).
\end{equation}
If
\begin{equation}
    \varphi(v,u)=\exp\!\left(\int_u^v a(s)\,\dd s\right),
\end{equation}
variation of constants weights $p(u+s)$ by
$\varphi(u+h,u+s)b(u+s)$.  Normalizing this weight gives the corresponding
step-average prediction whenever its integral is nonzero.  ELF uses
$u=-\log(1-t)$ on finite active intervals; the LangFlow update after dividing
by the signal coefficient uses $u=-\gamma/2$ and
$h=(\gamma_t-\gamma_s)/2$.  The ELF terminal endpoint has an infinite
unregularized log-clock width; the sampler uses the bounded interval
feature defined in Appendix~\ref{app:history_details} for this step.

\subsection{Correction for a finite, zero-initialized EMA}
\label{app:ema_identity}

For an affine sequence in the model-evaluation index, $p_i=p_0+iv$, with
$E_{i+1}=(1-\alpha)E_i+\alpha p_i$, $E_0=0$, and
$\alpha\in(0,1]$, let $r=1-\alpha$.  Direct summation gives
$w_np_n-E_n=L_nv$, where $w_n=1-r^n$ and
\begin{equation}
    L_n=\alpha\sum_{j=1}^{n}j\,r^{j-1}
       =\frac{1-(n+1)r^n+nr^{n+1}}{\alpha}.
    \label{eq:ema_lag}
\end{equation}
The raw residual satisfies $p_n-E_n=r^np_n+L_nv$, so the
bias-corrected increment $\widehat v=(w_np_n-E_n)/L_n$ recovers $v$
exactly.  For $\alpha\in(0,1)$ and $n\ge1$, the lag $L_n$ is
strictly positive and increases monotonically toward $1/\alpha$; at
$\alpha=1$, $L_n=1$.  This identity concerns increments per model evaluation.
Converting them to derivatives in solver time also requires the time-grid
spacing and prediction curvature.  The sampler uses the raw residual as a
history-based direction.

\section{Method Details}
\label{app:method_details}

\subsection{Principal-angle directions for self-conditioning}
\label{app:recurrent_transform}

Let $W_z$ and $W_m$ be the effective frozen input-projection matrices for the
latent and self-conditioning inputs.  Orthonormal row-space bases define
projectors $P_z$ and $P_m$.  On the self-conditioning row space,
\begin{equation}
    P_mP_zP_mQ
      =Q\diag(\rho_1^2,\ldots,\rho_r^2),
    \qquad 0\leq\rho_i\leq1.
    \label{eq:canonical_overlap}
\end{equation}
The basis $Q$ and squared cosines $\rho_i^2$ are computed once from frozen
model weights.

At step~$k$, the transform uses
\begin{equation}
    \widehat a_k=\kappa_{k-1}(1-\bar\delta_{k-1}),
\end{equation}
where $\kappa_{k-1}$ is the solver's update weight from~\eqref{eq:solver_weighted_state}
and $\bar\delta_{k-1}$ is the average clipped cosine distance defined in
Appendix~\ref{app:empirical_carrier_statistics}.  The state initialization
$\kappa_{-1}=0$, $\bar\delta_{-1}=0$ makes $F_0=I$.
Let $\lambda_R^{(0)}\ge0$ be the fixed base transform exponent,
$\lambda_R=\lambda_R^{(0)}\eta_R$, $r\ge1$, and
$0<\epsilon_{\rm fp}\le1$.  Then
\begin{equation}
\begin{aligned}
    b_{k,i}&=\max\{1-\widehat a_k\rho_i^2,\epsilon_{\rm fp}\},\\
    g_k&=\left(\prod_{i=1}^{r}b_{k,i}\right)^{1/r},\\
    \ell_{k,i}&=(g_kb_{k,i})^{\lambda_R}.
\end{aligned}
    \label{eq:released_retention}
\end{equation}
The resulting transform is
\begin{equation}
    F_k=I-Q(I-\diag(\ell_{k,1},\ldots,\ell_{k,r}))Q^\top.
\end{equation}
It is symmetric, preserves the orthogonal complement of $Q$, and multiplies
principal direction $i$ by $\ell_{k,i}$.  The floor $\epsilon_{\rm fp}$ prevents complete suppression of any
direction.  The common factor $g_k^{\lambda_R}$ adds a uniform
contraction within the principal subspace; because
$0<b_{k,i}\le1$ and $b_{k,i}$ is nonincreasing in $\rho_i^2$, the
transform satisfies the contraction conditions in
Section~\ref{sec:conditional_innovation}.  Setting this factor to one isolates the direction-dependent
dampening in the experiment reported in
Appendix~\ref{app:reference_experiments}.

\subsection{Input-projection geometry across models}
\label{app:projection_geometry}

Table~\ref{tab:projection_geometry} reports the effective input-projection
matrices used to compute the principal-angle basis.
For ELF, the latent and self-conditioning inputs pass through a shared
bottleneck projection before the first transformer block, producing
effective matrices of shape $128\times512$ (rank~$128$ in a
$512$-dimensional input space).  Because the row spaces are
$128$-dimensional subspaces of~$\mathbb R^{512}$, the principal-angle
spectrum is non-trivial: squared cosines range from near zero to~$0.92$
with a well-spread distribution that is stable across checkpoints.

For LangFlow, the self-conditioning projection is $768\times1536$ and
splits into two $768\times768$ blocks.  Both are full rank, so
$P_z{=}P_m{=}I$ and all squared cosines equal one.  The self-conditioning
transform therefore applies isotropic dampening in this architecture.

\begin{table}[H]
\centering
\caption{Effective input-projection geometry.  ELF projections pass
through a $128$-wide bottleneck, producing a well-spread principal-angle
spectrum; LangFlow projections are full rank and the transform applies
isotropic dampening.  Canonical correlations $\rho_i$ are the cosines of
the principal angles between the two row spaces.}
\label{tab:projection_geometry}
\small
\setlength{\tabcolsep}{4.2pt}
\begin{tabular}{lcccccccc}
\toprule
Model & Shape & Rank & $\rho_{\min}$ & $\rho_{\rm med}$ & $\rho_{\max}$
 & Mean $\rho_i^2$ & $|\rho_i{\ge}0.9|$ & $|\rho_i{\le}0.1|$ \\
\midrule
ELF-B & $128{\times}512$ & 128 & 0.001 & 0.594 & 0.959
 & 0.391 & 18 & 14 \\
ELF-M & $128{\times}512$ & 128 & 0.001 & 0.581 & 0.950
 & 0.381 & 17 & 13 \\
ELF-L & $128{\times}512$ & 128 & 0.004 & 0.577 & 0.952
 & 0.375 & 13 & 13 \\
\midrule
LangFlow & $768{\times}768$ & 768 & \multicolumn{5}{c}{full rank;
all $\rho_i{=}1$} \\
\bottomrule
\end{tabular}
\end{table}

\subsection{Prediction history and solver-step width}
\label{app:history_details}

The EMA uses weight $\alpha\in(0,1]$ on the current prediction, with
$E_0=0$.  The history residual used by the sampler is
\begin{equation}
    H_k=\mathbf{1}[\sigma_k>0](p_k-E_k),
    \label{eq:history_direction}
\end{equation}
and is active on steps with nonzero precomputed statistics.  The
zero-initialized EMA introduces a startup bias that decays as
$(1{-}\alpha)^k$; a bias-corrected variant
$\widehat v_k{=}(w_kp_k{-}E_k)/L_k$%
\phantomsection\label{eq:finite_history_increment}%
\ exists (Appendix~\ref{app:ema_identity}), but the sampler uses the raw
residual since the bias vanishes after a few steps.

A second direction estimate compares $p_k$ with the running weighted
average~$O_k$ and is rescaled samplewise to match $\|H_k\|_F$:
\begin{equation}
    G_k=\begin{cases}\norm{H_k}_F
        \dfrac{p_k-O_k}{\norm{p_k-O_k}_F}&\text{if }\|H_k\|_F,\|p_k-O_k\|_F>\epsilon,\\[4pt]0&\text{otherwise,}\end{cases}
    \label{eq:solver_history_direction}
\end{equation}
as in \eqref{eq:direction_estimates}.
The history-based correction direction is
\begin{equation}
    U_k=H_k+\eta_O\,\Delta u_k\,(H_k-G_k).
    \label{eq:released_transport_estimator}
\end{equation}
It reduces to $H_k$ when the two estimates agree or the step width vanishes.
For ELF, let $t_k^{\rm start}$ be the official solver-grid start before any
re-noising and define
\begin{equation}
c_\epsilon(t)=
\begin{cases}
-\log(1-t),&t\le1-\epsilon,\\
-\log\epsilon+\dfrac{t-(1-\epsilon)}{\epsilon},&t>1-\epsilon,
\end{cases}
\qquad \widetilde c_\epsilon(t)=\frac{c_\epsilon(t)}{c_\epsilon(1)}.
\end{equation}
The feature used by the sampler is
$\Delta u_k=\widetilde c_\epsilon(t_{k+1})-
\widetilde c_\epsilon(t_k^{\rm start})$.  LangFlow uses
$\Delta u_k=(\gamma_k-\gamma_{k+1})/(\gamma_{\max}-\gamma_{\min})$.
Both lie in $[0,1]$ on their sampling grids.  The exact solver-time widths are
$-\log(1-t_{k+1})+\log(1-s_k)$ and
$(\gamma_k-\gamma_{k+1})/2$, respectively
(Appendix~\ref{app:interval_proof}).

\subsection{Empirical statistics from reference corrections}
\label{app:empirical_carrier_statistics}

The offline statistics are computed from a fixed reference recurrence
that applies a correction during the middle portion of sampling and
ramps smoothly.  The specific constants below define this reference
and are not varied per configuration.
Let $s_k^{\rm ref}$ be a step-dependent activation schedule,
$g_T^{(0)}$ a fixed base scale, and
$\lambda_k^{\rm ref}=3.5\,g_T^{(0)}s_k^{\rm ref}$.  On each reference
trajectory, the correction is
\begin{equation}
\begin{aligned}
\widetilde H_k&=\lambda_k^{\rm ref}(p_k-E_k),\\
\widetilde G_k&=
\begin{cases}
\|\widetilde H_k\|_F\dfrac{p_k-O_k}{\|p_k-O_k\|_F},
 &\|\widetilde H_k\|_F,\|p_k-O_k\|_F>\epsilon,\\[3pt]
0,&\text{otherwise},
\end{cases}\\
\widetilde U_k&=\widetilde H_k+
\Delta u_k\,\widetilde\delta_k\odot
(\widetilde H_k-\widetilde G_k).
\end{aligned}
\label{eq:reference_carrier}
\end{equation}
The activation schedule is
$s_k^{\rm ref}=\mathbf 1[0.40<r_k\le0.95]
\sqrt{(r_k-0.40)/0.55}$ for normalized sampling progress~$r_k$; inactive
tokens are zeroed throughout.  During generation,
\Eqref{eq:released_transport_estimator} uses $H_k=p_k-E_k$ on every step
with $\sigma_k>0$.
For active token $n$, $\widetilde\delta_{k,n}$ is
$\min\{1,1-\cos(\widetilde H_{k,n},\widetilde G_{k,n})\}$ when both vectors
are nondegenerate and zero otherwise.  The reference recurrence supplies
$p_k+\widetilde U_k$ to the solver and stores the raw $p_k$ for the next
self-conditioning input.

For a correction tensor $V$ on active positions $I$, let
$\mathsf M_I(V)$ be its token mean and
$C_I V=V-\mathsf M_I(V)$ its centered component.  Two disjoint frozen
trajectory banks $\mathcal B_\mu$ and $\mathcal B_\sigma$ define
\begin{equation}
\begin{aligned}
\bar\delta_k
 &=\mathbb E_{\xi\sim\mathcal B_\mu}
   \left[\frac1{|I_k^\xi|}\sum_{n\in I_k^\xi}
   \widetilde\delta_{k,n}^\xi\right],\\
\mu_k
 &=\mathbb E_{\xi\sim\mathcal B_\mu}
   [\mathsf M_{I_k^\xi}(\widetilde U_k^\xi)],\\
\sigma_k^2
 &=\frac{\sum_{\xi\in\mathcal B_\sigma}
          \|C_{I_k^\xi}\widetilde U_k^\xi\|_F^2}
         {\sum_{\xi\in\mathcal B_\sigma}|I_k^\xi|D},
\end{aligned}
\label{eq:empirical_carrier_statistics}
\end{equation}
where $D$ is the prediction width.  Thus $\mu_k$ and $\sigma_k$ summarize the
correction used by the fixed reference recurrence.  The true correction
$C_k^*$ enters separately through the endpoint condition in
\eqref{eq:positive_alignment_condition}.  As with empirical model
statistics in training-free solvers~\citep{dpm_solver_v3}, the update form is
fixed analytically and the frozen model and grid supply its coefficients.
Here $\mathbb E_{\xi\sim\mathcal B}$ denotes a uniform empirical average over
the stored bank.

\subsection{Matching the empirical mean and RMS}
\label{app:correction_details}

For active positions $I_k$, let $C_{\rm tok}$ center over $I_k$ and
$\RMS_{I_k}$ compute the root-mean-square over active positions and width.
If $\RMS(C_{\rm tok}U_k)>0$, the unscaled projection $\Pi_k(U_k)$ in
\eqref{eq:released_solver_correction} satisfies
\begin{equation}
 \mathsf M_{I_k}(\Pi_k)=\mu_k,\qquad
 \RMS_{I_k}(C_{\rm tok}\Pi_k)=\sigma_k.
\end{equation}
Among all tensors sharing these two moments, the centered component
uniquely maximizes alignment with $C_{\rm tok}U_k$ (by Cauchy--Schwarz).
\phantomsection\label{prop:empirical_projection}
The outer strength $\eta_T$ scales both moments, so the final correction $A_k$
has mean $\eta_T\mu_k$ and centered RMS $\eta_T\sigma_k$.  Inactive positions
receive zero correction.  If the centered direction has zero norm, the
implementation retains only $\eta_T\mu_k$.

The affine endpoint condition in~\eqref{eq:positive_alignment_condition}
now applies with $B_k=\Pi_k(U_k)$.  Whenever
$\ip{C_k^*}{\Pi_k(U_k)}>0$, a nonempty interval of positive
$\eta_T$ strictly improves the endpoint obtained from~$p_k$ alone.

To separate the two contributions, write $N_k=|I_k|$ and
$\overline C_k^*=\mathsf M_{I_k}(C_k^*)$.  When the centered target is nonzero,
$\theta_k$ denotes the angle between $C_{\rm tok}C_k^*$ and
$C_{\rm tok}U_k$.  Orthogonality of the token-mean and centered components gives
the exact decomposition
\begin{equation}
\begin{aligned}
\ip{C_k^*}{\Pi_k(U_k)}_F
={}&N_k\ip{\overline C_k^*}{\mu_k}
 +\sigma_k\frac{\ip{C_{\rm tok}C_k^*}{C_{\rm tok}U_k}_F}
 {\RMS_{I_k}(C_{\rm tok}U_k)}\\
={}&N_k\ip{\overline C_k^*}{\mu_k}
 +\sigma_k\sqrt{N_kD}\,
   \|C_{\rm tok}C_k^*\|_F\cos\theta_k.
\end{aligned}
\label{eq:alignment_decomposition}
\end{equation}
The first term depends on how well the precomputed mean~$\mu_k$ aligns
with the true correction's mean; the second depends on the
angle~$\theta_k$ between the centered correction direction and the
centered target.  Frozen reference trajectories set $\mu_k$
and~$\sigma_k$; the current trajectory determines $\theta_k$
through~$U_k$.  The endpoint is strictly improved whenever
$2\ip{C_k^*}{\Pi_k(U_k)}_F>
\eta_T\|\Pi_k(U_k)\|_F^2$.

\section{Additional Experimental Details and Results}
\label{app:experiment_details}

\subsection{Evaluation protocol}
\label{app:evaluation_protocol}

Every matched comparison uses the same frozen checkpoint, initial latent,
base solver, time grid, sequence length, and number of model evaluations.
Each configuration starts from the same random stream; generation and
evaluation are run separately to avoid interference.

ELF uses its official random logit-normal grid.  Its SDE noise scale is
$2.0$, $2.0$, $1.5$, and $1.0$ at $8$, $16$, $32$, and $64$~NFE.
The official self-conditioning scale is $3.0$ for all evaluated NFE.
Evaluation uses four ranks and official seeds $0$--$5$; each seed produces
1,024 generations of length 1,024.  The independent seed run is the unit for
the reported standard error.

LangFlow uses its official time grid, obtained from quantiles of the proposal
distribution between $1-10^{-5}$ and $10^{-5}$, and seeds $0$--$5$.  Each
seed produces 1,024 generations per setting.  OpenWebText samples have
length 1,024 and LM1B samples have length 128.

ELF computes generative perplexity with frozen GPT-2 Large after applying its
official filter to remove empty decoded samples, then retokenizes decoded text
for unigram entropy.
LangFlow computes entropy from generated token IDs and perplexity after
decoding.  Official and \method{} configurations use the same evaluator convention in
each comparison.

Our ELF integration reproduces the official sampling code exactly:
under matched model weights and random streams, all intermediate and
final tensors agree elementwise across ranks.  The LangFlow
integration is verified analogously on closed-loop generation outputs.

\subsection{Pairwise LLM-judge evaluation protocol}
\label{app:arena_protocol}

The pairwise evaluation adapts the Arena-Hard-Auto~v2
protocol~\citep{arena_hard_auto} for unconditional text generation.
Every NFE in $\{8,16,32,64\}$ produces 1,024 text pairs by matching
official and \method{} generations on the same latent and sample index
(LangFlow OpenWebText, seed~42, 1,024-token sequences).  Each pair is
presented twice to the judge with the answer order exactly reversed,
yielding 8,192 judgments.

The judge (DeepSeek~V4~Pro, thinking enabled) receives neither method
names nor NFE labels.  It scores grammatical
fluency, discourse coherence, semantic content, and degeneration artifacts,
then returns a five-level verdict:
$A\!\gg\!B$, $A\!>\!B$, $A\!=\!B$, $B\!>\!A$, $B\!\gg\!A$.  Significant
verdicts ($\gg$) contribute three binary observations to the Arena score;
slight verdicts ($>$) and ties contribute one.

Pair-level outcomes merge the two games per pair: a pair is a candidate win
if the weighted score exceeds $0.5$, an official win if below $0.5$, and a
tie otherwise.  The 95\% confidence interval uses 10,000 paired bootstrap
resamples.  No judgments were excluded or reassigned.

\subsection{Complete endpoint results}
\label{app:complete_endpoint_results}

Tables~\ref{tab:elf_main} and~\ref{tab:langflow_main} give the absolute GenPPL,
uncertainty, and unigram entropy values corresponding to
Figure~\ref{fig:main_results}.

\begin{table}[H]
\centering
\caption{\method{} reduces GenPPL at every ELF NFE; the gap is largest at 8~NFE. Entries are mean $\pm$ standard error over six seeds, each with 1,024 samples.}
\label{tab:elf_main}
\small
\setlength{\tabcolsep}{5.2pt}
\begin{tabular}{llrrrr}
\toprule
&& \multicolumn{2}{c}{GenPPL $\downarrow$} & \multicolumn{2}{c}{$H$} \\
\cmidrule(lr){3-4}\cmidrule(lr){5-6}
Model & NFE & Official & \methodshort{} & Official & \methodshort{} \\
\midrule
\multirow{4}{*}{ELF-B}
& 8  & $70.6{\pm}1.7$ & $42.7{\pm}2.3$ & $5.23{\pm}.02$ & $5.10{\pm}.01$ \\
& 16 & $32.2{\pm}0.8$ & $28.7{\pm}1.0$ & $5.18{\pm}.01$ & $5.16{\pm}.01$ \\
& 32 & $23.7{\pm}0.3$ & $21.0{\pm}0.2$ & $5.15{\pm}.01$ & $5.14{\pm}.01$ \\
& 64 & $19.0{\pm}0.2$ & $18.0{\pm}0.3$ & $5.08{\pm}.01$ & $5.10{\pm}.01$ \\
\midrule
\multirow{4}{*}{ELF-M}
& 8  & $108.5{\pm}7.4$ & $52.3{\pm}3.5$ & $5.32{\pm}.03$ & $5.23{\pm}.00$ \\
& 16 & $38.7{\pm}1.4$ & $32.6{\pm}1.5$ & $5.31{\pm}.01$ & $5.29{\pm}.01$ \\
& 32 & $26.1{\pm}0.2$ & $20.5{\pm}0.2$ & $5.24{\pm}.00$ & $5.21{\pm}.00$ \\
& 64 & $21.7{\pm}0.2$ & $18.4{\pm}0.1$ & $5.18{\pm}.01$ & $5.20{\pm}.00$ \\
\midrule
\multirow{4}{*}{ELF-L}
& 8  & $104.9{\pm}5.3$ & $55.7{\pm}3.2$ & $5.31{\pm}.10$ & $5.08{\pm}.12$ \\
& 16 & $46.9{\pm}1.7$ & $41.9{\pm}2.1$ & $5.40{\pm}.01$ & $5.37{\pm}.01$ \\
& 32 & $30.1{\pm}0.3$ & $23.4{\pm}0.3$ & $5.34{\pm}.00$ & $5.29{\pm}.01$ \\
& 64 & $23.6{\pm}0.2$ & $19.1{\pm}0.1$ & $5.28{\pm}.00$ & $5.26{\pm}.00$ \\
\bottomrule
\end{tabular}
\end{table}

\begin{table}[H]
\centering
\caption{\method{} improves LangFlow quality at every NFE on both datasets, reducing OpenWebText GenPPL from $531$ to~$62$ at 8~NFE. Entries are mean $\pm$ standard error over six seeds, each with 1,024 samples (OpenWebText: 1,024 tokens; LM1B: 128 tokens).}
\label{tab:langflow_main}
\small
\setlength{\tabcolsep}{5.2pt}
\begin{tabular}{llrrrr}
\toprule
&& \multicolumn{2}{c}{GenPPL $\downarrow$} & \multicolumn{2}{c}{$H$} \\
\cmidrule(lr){3-4}\cmidrule(lr){5-6}
Data & NFE & Official & \methodshort{} & Official & \methodshort{} \\
\midrule
\multirow{6}{*}{OWT}
& 8   & $531.2{\pm}3.4$ & $61.6{\pm}0.2$ & $5.81{\pm}.00$ & $5.33{\pm}.00$ \\
& 16  & $234.5{\pm}0.8$ & $48.8{\pm}0.1$ & $5.69{\pm}.00$ & $5.30{\pm}.00$ \\
& 32  & $123.0{\pm}0.5$ & $43.8{\pm}0.1$ & $5.58{\pm}.00$ & $5.31{\pm}.00$ \\
& 64  & $79.3{\pm}0.3$  & $40.0{\pm}0.1$ & $5.50{\pm}.00$ & $5.31{\pm}.00$ \\
& 128 & $59.4{\pm}0.1$  & $38.3{\pm}0.1$ & $5.43{\pm}.00$ & $5.31{\pm}.00$ \\
& 256 & $48.6{\pm}0.2$  & $39.6{\pm}0.1$ & $5.36{\pm}.00$ & $5.31{\pm}.00$ \\
\midrule
\multirow{6}{*}{LM1B}
& 8   & $264.0{\pm}1.3$ & $73.0{\pm}0.1$ & $4.38{\pm}.00$ & $4.24{\pm}.00$ \\
& 16  & $157.1{\pm}0.6$ & $63.2{\pm}0.2$ & $4.35{\pm}.00$ & $4.24{\pm}.00$ \\
& 32  & $113.8{\pm}0.7$ & $60.8{\pm}0.1$ & $4.33{\pm}.00$ & $4.26{\pm}.00$ \\
& 64  & $94.3{\pm}0.3$  & $61.4{\pm}0.1$ & $4.32{\pm}.00$ & $4.27{\pm}.00$ \\
& 128 & $84.4{\pm}0.1$  & $61.9{\pm}0.2$ & $4.31{\pm}.00$ & $4.28{\pm}.00$ \\
& 256 & $79.3{\pm}0.2$  & $63.0{\pm}0.2$ & $4.31{\pm}.00$ & $4.29{\pm}.00$ \\
\bottomrule
\end{tabular}
\end{table}

\begin{table}[H]
\centering
\caption{Word-level 4-gram repetition rate (word Rep-4) for LangFlow
endpoints.  Entries are mean $\pm$ standard error over six seeds.}
\label{tab:repetition}
\small
\setlength{\tabcolsep}{5.2pt}
\begin{tabular}{llrr}
\toprule
&& \multicolumn{2}{c}{Word Rep-4 ($\times10^{-3}$)} \\
\cmidrule(lr){3-4}
Data & NFE & Official & \methodshort{} \\
\midrule
\multirow{6}{*}{OWT}
& 8   & $0.11{\pm}.01$ & $1.82{\pm}.04$ \\
& 16  & $0.51{\pm}.02$ & $3.59{\pm}.10$ \\
& 32  & $1.54{\pm}.06$ & $5.33{\pm}.16$ \\
& 64  & $3.95{\pm}.08$ & $7.78{\pm}.19$ \\
& 128 & $8.38{\pm}.19$ & $11.92{\pm}.31$ \\
& 256 & $16.91{\pm}.41$ & $18.51{\pm}.53$ \\
\midrule
\multirow{6}{*}{LM1B}
& 8   & $0.07{\pm}.01$ & $0.47{\pm}.02$ \\
& 16  & $0.18{\pm}.02$ & $0.67{\pm}.04$ \\
& 32  & $0.34{\pm}.04$ & $0.66{\pm}.05$ \\
& 64  & $0.57{\pm}.05$ & $0.87{\pm}.06$ \\
& 128 & $0.81{\pm}.05$ & $1.11{\pm}.07$ \\
& 256 & $1.11{\pm}.10$ & $1.34{\pm}.09$ \\
\bottomrule
\end{tabular}
\end{table}

\FloatBarrier

\subsection{Precomputation and inference-time settings}
\label{app:precomputation_details}

The principal-angle directions for the latent and self-conditioning input
projections are computed once from the frozen input weights.  The per-step
statistics in this subsection use only latent and prediction tensors.
For ELF, finite differences measure
the first-order model responses to perturbing the self-conditioning input and
the prediction supplied to the solver.  Their $2\times2$ Gram matrix determines
the base transform exponent $\lambda_R^{(0)}$ and the reference correction
scale $g_T^{(0)}$.  For LangFlow, the full-rank projection geometry
(Appendix~\ref{app:projection_geometry}) gives
$\lambda_R^{(0)}=g_T^{(0)}=1$; the coefficients in
Table~\ref{tab:strengths} are effective strengths directly.
For ELF, denote the response Gram matrix by~$G$, the total strength by~$s_0$,
and the relative solver weight by~$\varphi$.  Define
$v=(1-\varphi,\varphi)^\top$ and $\mathbf 1=(1,1)^\top$.  The resulting scales
are
\begin{equation}
 \chi=\sqrt{\frac{\mathbf 1^\top G\mathbf 1}{v^\top Gv}},\qquad
 \lambda_R^{(0)}=s_0\chi(1-\varphi),\qquad
 g_T^{(0)}=s_0\chi\varphi.
\end{equation}
We set $s_0=\varphi=0.75$, fixing the relative first-order
response norm of the two interventions.

The empirical per-step statistics in
\eqref{eq:empirical_carrier_statistics} use independent trajectory banks.
ELF combines four sets of 1,024 trajectories for each model and NFE.  Disjoint
halves estimate the token mean and average clipped cosine distance on one
side and the centered RMS on the other.  LangFlow uses 128 independent latents
with a disjoint $64/64$ split.  These quantities summarize the correction in
the fixed reference recurrence.  High-resolution step averages are used only
in the separate diagnostics of Appendix~\ref{app:reference_experiments}.

Table~\ref{tab:strengths} lists the inference-time coefficients for all
evaluated configurations.  Lower NFE generally uses stronger
correction, as fewer steps amplify the solver-introduced coupling
(Section~\ref{sec:coupling_experiment}).
On a single H100 GPU, the offline precomputation for one ELF-B NFE
takes 128 seconds at 8~NFE and 401 seconds at 64~NFE.  LangFlow precomputes
all evaluated NFE in a single pass of 496 seconds (OWT) and
180 seconds (LM1B).

\begin{table}[t]
\centering
\caption{Inference-time coefficients (Algorithm~\ref{alg:method}).
``---'' marks the base value~$1$ (unchanged).}
\label{tab:strengths}
\small
\setlength{\tabcolsep}{4.6pt}
\begin{tabular}{lrcccc}
\toprule
System & NFE & $\eta_R$ & $\eta_T$ & $\eta_O$ & $\alpha$ \\
\midrule
\multirow{4}{*}{ELF-B}
& 8  & 3.3 & 1.7 & --- & 0.7 \\
& 16 & 2.3 & 0.5 & --- & 0.8 \\
& 32 & 2.3 & 1.1 & --- & 0.8 \\
& 64 & 1.5 & --- & 1.5 & 0.8 \\
\midrule
\multirow{4}{*}{ELF-M}
& 8  & 2.3 & 1.5 & 6.0 & 0.8 \\
& 16 & 1.3 & 0.9 & --- & 0.8 \\
& 32 & 1.7 & 1.7 & 0.7 & --- \\
& 64 & 3.3 & 2.5 & 0.8 & 0.8 \\
\midrule
\multirow{4}{*}{ELF-L}
& 8  & 2.3 & 2.0 & 2.0 & 0.5 \\
& 16 & 3.0 & 0.8 & 1.5 & 0.8 \\
& 32 & 2.5 & 2.2 & --- & --- \\
& 64 & 3.3 & 3.3 & 0.5 & 0.8 \\
\midrule
\multirow{4}{*}{\shortstack[l]{LangFlow\\OWT}}
& 8        & 2.5 & 1.9 & --- & --- \\
& 16, 32   & 1.3 & 1.7 & --- & --- \\
& 64, 128  & 1.3 & 1.9 & --- & 0.8 \\
& 256      & 1.3 & 1.3 & --- & 0.8 \\
\midrule
\multirow{4}{*}{\shortstack[l]{LangFlow\\LM1B}}
& 8        & 2.5 & 2.5 & --- & --- \\
& 16       & 1.3 & 2.5 & --- & 0.9 \\
& 32, 128  & 0.6 & 2.5 & --- & 0.8 \\
& 64, 256  & 1.3 & 2.5 & --- & 0.8 \\
\bottomrule
\end{tabular}
\end{table}

\subsection{Changing correlation while preserving each noise-level distribution}
\label{app:coupling_protocol}

The coupling experiment uses 256 fresh OpenWebText examples with frozen ELF-B
at 8 and 16~NFE.  For each example and correlation in
$\{0,0.25,0.5,0.75,1\}$, it constructs successive standardized Gaussian
corruptions with the specified correlation.  This preserves the Gaussian
marginal at every evaluation.  Within each correlation setting, the
unmodified and modified model evaluations receive exactly the same current
latent, time, and noise.

Denote the base prediction by $v_{\rm base}$ and the prediction after applying
one fixed self-conditioning transform by $v_{\rm sc}$.  For the fixed
target~$y$, define $d=v_{\rm sc}-v_{\rm base}$ and
$e=v_{\rm base}-y$.  The reported coefficient is the exact empirical
quadratic-risk optimum
along this single direction,
\begin{equation}
 \alpha^*(\rho)=-\frac{\sum\ip{e}{d}}
                         {\sum\|d\|^2},
 \qquad v(\alpha)=v_{\rm base}+\alpha d.
\end{equation}
This coefficient measures interpolation along the fixed
self-conditioning-induced direction and is used only in this diagnostic.
Table~\ref{tab:coupling} reports the continuous optimum and the risk of
applying the modification with coefficient one.  The optimum decreases
monotonically at both NFE, while the same modification applied
with coefficient one crosses from lower to higher risk than the base model.
The paired $\rho=0$ minus $\rho=1$ differences in the optimum are $0.58$ at
8~NFE and $0.43$ at 16~NFE, with 95\% intervals $[0.55,0.62]$ and
$[0.40,0.46]$.

\begin{table}[t]
\centering
\caption{Changing temporal correlation while preserving the marginal
distribution at each noise level.  $\alpha^*$ is the risk-optimal interpolation
coefficient along one fixed self-conditioning-induced prediction direction;
the ratio compares prediction risk at coefficient one with the base model.}
\label{tab:coupling}
\small
\setlength{\tabcolsep}{5.2pt}
\begin{tabular}{rrrrr}
\toprule
& \multicolumn{2}{c}{8~NFE} & \multicolumn{2}{c}{16~NFE} \\
\cmidrule(lr){2-3}\cmidrule(lr){4-5}
$\rho$ & $\alpha^*$ & risk ratio & $\alpha^*$ & risk ratio \\
\midrule
0.00 & 0.88 & 0.92 & 0.78 & 0.92 \\
0.25 & 0.79 & 0.94 & 0.69 & 0.94 \\
0.50 & 0.62 & 0.97 & 0.55 & 0.98 \\
0.75 & 0.45 & 1.01 & 0.40 & 1.05 \\
1.00 & 0.30 & 1.07 & 0.35 & 1.11 \\
\bottomrule
\end{tabular}
\end{table}

Controls using a preliminary prediction computed from the current noisy state
have risk ratios near $1.02$;
controls without self-conditioning have ratios within $10^{-4}$ of~$1$.  The
quadratic risk decomposition is exact to within $10^{-8}$,
confirming that the reported coefficients are not affected by numerical
artifacts.

\subsection{Control transforms with the same eigenvalues}
\label{app:spectrum_protocol}

The experiment in Table~\ref{tab:spectrum_pairing} disables the solver
correction and uses 256 fresh ELF-B generations at 8~NFE.  Every configuration
supplies the raw prediction to the solver, stores it for the next model
evaluation, and makes eight model evaluations.  The overlap-aligned, reverse,
and random transforms have exactly the same set of eigenvalues and mean
strength.  Their only difference is the assignment of these eigenvalues to the
principal directions.

For a self-conditioning input $m$, define
\begin{equation}
    \mathcal E(m)=\frac12\sum_{n\in I_k}\sum_i
      \rho_i^2\ip{m_n}{q_i}^2,
    \label{eq:weighted_projection_energy}
\end{equation}
where $q_i$ is principal direction $i$.  Table~\ref{tab:spectrum_pairing}
reports the cosine similarity between the self-conditioning modification and
$-\nabla\mathcal E$, the decrease in $\mathcal E$ divided by the modification
RMS, and both endpoint metrics.  Table~\ref{tab:spectrum_full}
repeats the three eigenvalue assignments with the solver correction active.
The overlap-aligned assignment gives the highest cosine similarity, the largest normalized
decrease in $\mathcal E$, and the lowest two likelihood measures.  The configuration
with solver correction but no self-conditioning transform separates the
gain due to the transform in this matched comparison.

\begin{table}[t]
\centering
\caption{\textbf{The overlap-aligned transform outperforms matched
direction assignments.}  Every solver correction is disabled.
Overlap-aligned, reverse, and random use the same retention eigenvalues and
differ only in their assignment to principal directions.  Alignment is the
cosine with the steepest decrease in the overlap objective $\mathcal{E}$.}
\label{tab:spectrum_pairing}
\small
\setlength{\tabcolsep}{4.5pt}
\begin{tabular}{lrrr}
\toprule
Configuration & Alignment $\uparrow$ & $\Delta\mathcal{E}/\RMS$ $\downarrow$
 & GenPPL $\downarrow$ \\
\midrule
Official & -- & -- & 64.27 \\
Overlap-aligned & \textbf{0.94} & \textbf{$-40.35$k}
          & \textbf{55.64} \\
Reversed  & 0.51 & $-22.94$k & 73.95 \\
Random    & 0.74 & $-32.15$k & 59.59 \\
\bottomrule
\end{tabular}
\end{table}

\begin{table}[t]
\centering
\caption{Self-conditioning transforms with the solver correction active.
The three transformed configurations use the same eigenvalues, assigned to
principal directions in different orders.  $\mathcal E$ is defined in
\eqref{eq:overlap_energy}.}
\label{tab:spectrum_full}
\small
\setlength{\tabcolsep}{4.6pt}
\begin{tabular}{lrrr}
\toprule
Configuration & Alignment $\uparrow$ & $\Delta\mathcal E/\RMS$ $\downarrow$
 & GenPPL $\downarrow$ \\
\midrule
Official SC, no solver correction & -- & -- & 65.23 \\
Official SC, solver correction & -- & -- & 38.17 \\
Overlap-aligned & \textbf{0.94} & \textbf{$-53.62$k}
          & \textbf{34.93} \\
Reversed  & 0.54 & $-32.52$k & 40.21 \\
Random    & 0.75 & $-43.69$k & 36.92 \\
\bottomrule
\end{tabular}
\end{table}

\subsection{Reference averages over one solver step}
\label{app:interval_protocol}
\label{app:reference_experiments}

Subdividing each solver step into 32 substeps provides a high-resolution
reference for the step-average prediction.  Across 8, 16, and 32~NFE, the
correction $C_k^*$ scales as first order in step width (fitted exponent
$1.04$, 95\% CI $[1.02,1.05]$) and its contribution to the latent update
scales as second order (fitted exponent $2.02$, 95\% CI $[2.01,2.04]$),
matching the predicted orders from
Section~\ref{sec:interval_prediction}.  The EMA-based estimate from past
predictions outperforms equal-RMS permutations at every NFE.

The scaling experiment uses frozen ELF-B at 8, 16, and
32~NFE.  Every setting contains 128 fresh trajectories.  The first 64 fit
one nonnegative coefficient for each estimate derived from prediction history;
the remaining 64 evaluate normalized MSE and equal-RMS permuted controls.
Confidence intervals use 4,000 sample bootstrap resamples.

Table~\ref{tab:target_read_endpoint} isolates the self-conditioning
transform on 256 fresh ELF-B samples at 16~NFE with the solver correction
disabled.  Direction-dependent dampening outperforms both the official
transform and uniform scaling, confirming that aligning dampening with
the principal-angle spectrum improves generation quality.

\begin{table}[H]
\centering
\caption{Self-conditioning controls on 256 ELF-B samples at 16~NFE with the
solver correction disabled.  The overlap-aligned transform dampens
principal direction $i$ by $\max(1-\widehat a_k\rho_i^2,\epsilon_{\rm fp})$
before any global scaling.}
\label{tab:target_read_endpoint}
\small
\setlength{\tabcolsep}{4.8pt}
\begin{tabular}{lrrr}
\toprule
Configuration & GenPPL $\downarrow$ & GPT-2 NLL $\downarrow$
 & $H$ \\
\midrule
Official & 29.44 & 3.39 & 5.13 \\
Overlap-aligned dampening, no global scaling & \textbf{26.56} & \textbf{3.28}
  & 5.06 \\
Overlap-aligned dampening, method scaling & 27.42 & 3.32 & 5.08 \\
Uniform scaling & 29.90 & 3.40 & 5.13 \\
\bottomrule
\end{tabular}
\end{table}

\subsection{Where to apply the solver correction across checkpoints}
\label{app:state_assignment_generalization}

The main experiment on where to apply the solver correction uses 1,024 matched
ELF-B generations at 8~NFE.  A second experiment applies the same fixed solver
correction to five additional model and NFE settings with 64 samples
each and no parameter changes.  As shown in
Table~\ref{tab:state_generalization}, applying the
correction in the solver outperforms the official sampler on both
paired likelihoods.  Adding the same correction to the next self-conditioning
input underperforms the solver assignment.

\begin{table}[H]
\centering
\caption{Where to apply the solver correction across checkpoints.  The first
two differences apply it only in the solver update and compare with the
official sampler.  The last also adds it to the next self-conditioning input
and compares with applying it only in the solver.}
\label{tab:state_generalization}
\small
\setlength{\tabcolsep}{4.8pt}
\begin{tabular}{lrrrr}
\toprule
Model & NFE & GPT-2 $\Delta_{\rm solver}$ & OWT $\Delta_{\rm solver}$
 & OWT $\Delta_{\rm next\,SC}$ \\
\midrule
ELF-B & 16 & $-0.07$ & $-0.16$ & $+3.00$ \\
ELF-M & 8  & $-0.38$ & $-0.31$ & $+2.51$ \\
ELF-M & 16 & $-0.17$ & $-0.18$ & $+3.45$ \\
ELF-L & 8  & $-0.27$ & $-0.25$ & $+1.63$ \\
ELF-L & 16 & $-0.14$ & $-0.14$ & $+2.75$ \\
\bottomrule
\end{tabular}
\end{table}

\subsection{Additional ablations}
\label{app:additional_ablations}

The remaining tables use ELF-B, 8~NFE, the official SDE noise scale $2.0$,
official seed 0, and 1,024 matched generations per configuration.

\begin{table}[H]
\centering
\caption{Controls for the solver correction on ELF-B with eight model evaluations.}
\label{tab:direction_controls}
\small
\begin{tabular}{lrr}
\toprule
Solver correction & GenPPL $\downarrow$ & $H$ \\
\midrule
History-based direction with mean/RMS scaling & \textbf{45.63} & 5.14 \\
Common component only & 57.21 & 4.98 \\
History-based direction permuted across samples & 49.27 & 4.99 \\
Random direction with equal RMS & 71.45 & 5.02 \\
Official (no correction) & 74.18 & 5.27 \\
\bottomrule
\end{tabular}
\end{table}

Table~\ref{tab:direction_controls} changes the correction direction.
A random equal-RMS direction nearly recovers official GenPPL, confirming
that the correction gain requires the specific history-based direction.
Removing the sample-specific component raises GenPPL, and permuting it
across samples reduces the gain further.

\begin{table}[t]
\centering
\caption{Matched sampler comparison on frozen ELF-B at 8~NFE with 1,024
matched generations.  All methods use the same model, time grid, and
initial latent.  Momentum~\citep{diffusion_momentum},
HiGS~\citep{higs}, and ACE~\citep{ace} are described in
Section~\ref{sec:related_work}.}
\label{tab:sampler_comparison}
\small
\begin{tabular*}{0.50\linewidth}{@{\extracolsep{\fill}}lrr@{}}
\toprule
Sampler & GenPPL $\downarrow$ & $H$ \\
\midrule
Official & 74.18 & 5.27 \\
Momentum & 74.72 & 5.28 \\
ACE & 115.43 & 5.43 \\
HiGS & 56.90 & 5.18 \\
HiGS + ACE & 65.84 & 5.24 \\
\method{} & \textbf{45.63} & 5.14 \\
\bottomrule
\end{tabular*}
\end{table}

Table~\ref{tab:sampler_comparison} compares \method{} with training-free
sampling methods under matched conditions on ELF-B at 8~NFE\@.
\method{} reaches $45.63$ GenPPL versus $56.90$ for
HiGS~\citep{higs} and $74.18$ for the official sampler.
Combining HiGS with ACE~\citep{ace} yields $65.84$, above \method{}'s
solver correction alone ($50.04$; Table~\ref{tab:placement}b).
The paired improvement over HiGS with ACE is $0.36$ GPT-2 NLL, with
95\% interval $[0.35,0.38]$.

\subsection{Throughput}
\label{app:throughput}

Throughput is measured with the same compiled model, batch size, number of
model evaluations, and generated outputs for the official and modified
sampling loops.  Both loops are deterministic and produce outputs
identical to those used in the generation experiments.
Table~\ref{tab:throughput} reports total samples per second and their ratio.
The ratio ranges from $0.91$ to $0.98$, with geometric mean $0.96$:
the sampler retains over 95\% of the baseline throughput without requiring
additional model evaluations, since all auxiliary quantities are derived from
tensors already available in the sampling loop.

\begin{table}[H]
\centering
\caption{Sampling throughput under matched computation.  Official and
\method{} use the same compiled model, batch size, NFE, and
generated outputs.  Retained is method throughput divided by official
throughput.}
\label{tab:throughput}
\small
\setlength{\tabcolsep}{8pt}
\begin{tabular}{lrrrr}
\toprule
Model & NFE & Official (samples/s) & \method{} (samples/s) & Retained \\
\midrule
ELF-B & 8  & 72.98 & 67.59 & 92.6\% \\
      & 16 & 39.39 & 35.93 & 91.2\% \\
\midrule
ELF-M & 8  & 22.20 & 21.47 & 96.7\% \\
      & 16 & 11.78 & 11.38 & 96.6\% \\
\midrule
ELF-L & 8  & 10.42 & 10.21 & 98.0\% \\
      & 16 &  5.53 &  5.42 & 98.1\% \\
\midrule
\multicolumn{4}{r}{Geometric mean retained} & \textbf{95.5\%} \\
\bottomrule
\end{tabular}
\end{table}

\subsection{Parameter sensitivity}
\label{app:sensitivity}

\begin{figure}[H]
\centering
\includegraphics[width=\linewidth]{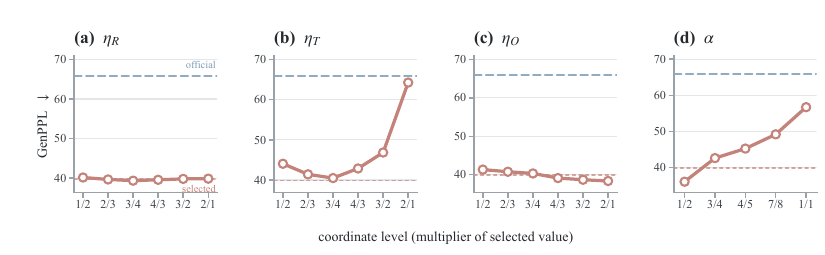}
\caption{One-at-a-time sensitivity of GenPPL to each coefficient on
ELF-B at 8~NFE.  Each panel varies one coefficient while holding the other
three at the values in Table~\ref{tab:strengths}.  The dashed
blue line marks the official sampler; the dotted red line marks the
Table~\ref{tab:strengths} value.
All 24 evaluated points reduce GenPPL relative to the official sampler.
The curves for $\eta_R$ and $\eta_O$ are nearly flat across the full
$[\times\!\frac{1}{2},\,\times\!2]$ range; doubling $\eta_T$ gives the highest
GenPPL among the tested variants.}
\label{fig:sensitivity}
\end{figure}

To assess coefficient robustness, we run a frozen
one-at-a-time sensitivity study on ELF-B at 8~NFE using 256 independent
paired samples that are disjoint from the reported evaluation data.  The
coefficient grid and evaluation protocol were fixed before any
generations were run; all 24 evaluated points are reported without
selection.

Every evaluated point improves GenPPL over the paired official sampler.
The self-conditioning coefficient $\eta_R$ varies GenPPL by less than 0.8 over
the full $[\times\!\frac{1}{2},\,\times\!2]$ range (39.4--40.2), and
$\eta_O$ varies by 3.0 (38.4--41.3).  The solver coefficient $\eta_T$ is
more sensitive: doubling it raises GenPPL to 64.3, still below the
official sampler's 65.8.  Reducing
$\eta_T$ to $\times\!\frac{1}{2}$ yields 44.0.  GenPPL increases monotonically
from 36.1 at $\alpha=\frac{1}{2}$ to 56.7 at $\alpha=1$.
Figure~\ref{fig:sensitivity} visualizes all four coefficient sweeps.

\subsection{Per-sample quality on LangFlow}
\label{app:sample_quality}

\begin{table}[H]
\centering
\caption{Paired-sample NLL comparison on LangFlow OpenWebText (six seeds,
1,024 samples per seed per NFE).  ``Higher NLL'' counts samples where
\method{} has higher GPT-2 NLL than the matched official sample.}
\label{tab:sample_failures}
\small
\setlength{\tabcolsep}{5pt}
\begin{tabular}{rrrrr}
\toprule
NFE & Paired samples & \methodshort{} higher NLL & Fraction & Mean $\Delta$NLL \\
\midrule
  8  & 6{,}144 &   0 & $0.00\%$ & $-2.16$ \\
 16  & 6{,}144 &   0 & $0.00\%$ & $-1.57$ \\
 32  & 6{,}144 &   2 & $0.03\%$ & $-1.03$ \\
 64  & 6{,}144 &  15 & $0.24\%$ & $-0.69$ \\
128  & 6{,}144 & 127 & $2.07\%$ & $-0.44$ \\
256  & 6{,}144 & 693 & $11.28\%$ & $-0.20$ \\
\bottomrule
\end{tabular}
\end{table}

Table~\ref{tab:sample_failures} reports per-sample NLL comparisons on
LangFlow OpenWebText.  At 8 and 16~NFE, every one of the 6,144 paired
samples has lower NLL under \method{}.

\subsection{Illustrative text comparisons}
\label{app:case_studies}

We reproduce one matched pair for each model family to illustrate the
decoded behavior.
LangFlow examples come from the reported six-seed generations; ELF examples
come from the disjoint 256-pair bank used in
Section~\ref{app:sensitivity}.  Each pair uses the same initial latent and
sample index.  The boxes contain the complete decoded generations, including
end-of-text markers when present.  Non-ASCII characters are transliterated
for typesetting; no span is truncated.

\begingroup
\raggedbottom

\paragraph{LangFlow--OpenWebText example.}
64~NFE, sample 599, $\Delta\mathrm{NLL}=-1.66$.  The \methodshort{}
generation removes the repeated phrase ``second place in second place'' and
sustains article-like segments for longer.
\begin{mdframed}[style=samplebox]
\footnotesize
\textbf{Official}\par\smallskip
\textless{}\textbar{}endoftext\textbar{}\textgreater{}

All of that in mind, of course, was that calling grand-peat--where Monaco start the Elite line at 
home--would improve to second place in second place overall in 2017. Ratings were spoiled from the 
first rankings, because Pabennis tried 19 points outrightly, and so far away to his best for 
competitive titles, but we quickly never predict the end of a steep, hot year within Giga Champions 
and one of this year's McFOR favourites.

So for traveling to Valencia the rest of the year--every, for example--that's my question of why, 
despite Sevan had so many promising moments in 2013, every player still felt the same cliche, 
wondrous pace about his own country, when in the Multiverse, it was Fernandez as a good roommate, 
not just my own wife and teenagers who had a dedicated, tactical approach.

For as long I went, the confidence I'd found throughout the entire season (that optimism happened 
now just as soon as I can imagine) was just real for the ideal year for anything. Valencia is 
there, we're ahead, and it still feels bad to see it coming yet quickly.\textless{}\textbar{}endoftext\textbar{}\textgreater{}DSP Spaces, 
the Jaaharabad Airways Ltd, is building the Qatar port and will connect it to Ahmedabad Airport for 
the cross-border services.

The port will begin at Mokintabad International Airport and will then connect to Panurbiba National 
Airport.

Although the airport will be closed, the port will also connect back to Disa, Ag-a-Palad and a 
public waychard known as Zone Na.

The Panurbiba Center will also serve the so-called Lid Ban log-airlets and outlets as part of the 
project.

The Lid Ban Center will also serve the conservation of water provided for the port's special water 
deck.

Besides the airport, DSP will move around the two-lane branches with an M/6 system and serve up to 
2 hundred passengers. The campus or 251 passengers.

Commenting at the project, DSP Chairman, Gindas Kumar, said the project would be "donural home for 
Ahmedabad people."

"With the Qatar port terminal, Zone Na will be Ahmedabad's major link at the border. All investment 
investments will be made into this airport to make a much smoother environment for Indians," said 
Kumar.

"The port will be extended from Zone Na to the Ahmedabad Air location without further delays for 
shuttle operators to no services from Ahmedabad International Airport. This will be a good haven 
for noncoperative passenger services and the port will have a fast link to its Qatar border."

Zone Na Director of visual Group Joan Engmave told reporters that "The port, the arch from Mecca to 
Danda, was handed over to Qatar, and the frame will soon be out."

Refaj Airport, Ahmedabad Airport and Matinadh Airport are the new Ahmedabad airports to take 
advantage of the Qatar proper. The project is another off-use airport such as the Public Satellite 
airport and welllands.\textless{}\textbar{}endoftext\textbar{}\textgreater{}Two weeks ago I waited for Nigel Farage's half-day residence 
near Queen Square in Goldstone Paul's, came on a passport "thatks our rise toward the independence 
vote" and sealed the decision to "born living Europeans that I didn't tell voters". "To assure you 
that this time round, or any other time of year, I must abide by the clear code of fact that I pass 
along for the visit," cracked alily. The passport had turned me over to the City. I stood 
confronting myself on Yorkshire Street, in Goldstone Paul's, suddenly mentioning Nigel Farage's MPs 
as if he had decided to call them "friends" in "Hab that Wh nearby". I was deeply informed of this 
distinction and certainly obliged to hypothesise about it from a source I would only suspect at 
Ukip if unspecified. It is still unsurprising, though, that the case of Ukip in general was derived 
from his previously made Southampton speculation. And there are some recent facts to examine.

But what Ukip's main milons, or the compassionate people themselves have told us about their spin, 
Nigel Farage's change implies an overt and strengthened leadership has almost nothing to do with 
people. Indeed it is true that many times net immigration left a devaluous stock of workers and 
brings in desperate people to reject it for access. The standards proposed by EU states concerning 
net immigration have made it clear that we should never let so many migrants to leave, because of 
the interests of the migrant and internally rather than the overseas taxpayers; which is precisely 
why it is common interest that Ukip's extreme leadership has then rightly sent us up into Europe. 
That past national day, in France, Mr Farage periodically holed himself and European 
countries\textless{}\textbar{}endoftext\textbar{}\textgreater{}

\par\medskip\textbf{\methodshort{}}\par\smallskip
\textless{}\textbar{}endoftext\textbar{}\textgreater{}

All of that in mind, of course, was a telling no-brainer, because Monaco won the Champions League 
in 2014 and slipped back to second place in the Champions League in 2017. Ratings were spoiled from 
the first round, because the Bavarian crashed 19 points outrightly, and so nobody seemed to know 
what for sure, because instead they struggled to predict the end of a very, big year within 
Borussia Dortmund and one of the world's top F1 runners.

So for me anyway, the out of the year (2016, for example) wasn't my question of why, the Bavarian 
had so many great moments in 2016, every league move, the same flirting, the aerrolic event, etc, 
but for the Bavarian, it was at last a great day, not just my own wife and children who had a 
great, great experience.

Despite how close I was, the passion I'd seen throughout the entire season (that has happened here 
just as often as I can imagine) was just personal for the sake of the thing. Monaco is special, 
they're strong, and it would be bad to see it come true...\textless{}\textbar{}endoftext\textbar{}\textgreater{}Hamid Karzai, the Haqqani 
Taliban headliner, indicated that the United States would have to bury Mr. Karzai for the 
cease-fire talks.

The new peace agreement, an apparent move by Mr. Karzai to meet with the United States, has been in 
force after the disputed elections of Sunday. The military has also engaged in an effort to 
preserve Mr. Karzai's authority holding hold as early as Fallujah.

But on the Islamic State's announcement, Mr. Hamid Karzai, president of the American Parliament, 
said the U.N. "absolutelyprised" that the United States would have to intervene.

Photo

Mr. Karzai added that the U.N. announcement about the renewed peace talks should allow the United 
States to take further steps and exchange its obligations as well as with the United Nations.

Advertisement Continue reading the main story

Mr. Karzai, who had called the peace talks "stalled" after Afghan fighters had rallied hard to take 
the coup push against Mr. Karzai, said he didn't know if the United States was trading in or if 
Karzai was able to bury the peace agreements just after carrying out the reconciliation.

Yol. Karzai signed off a deal to rein Talibanate Mr. Karzai's forces to keep Mr. Karzai in power. 
But Mr. Karzai has also formally cred the agreement and says he won't sign a final peace agreement.

Newsletter Sign Up Continue reading the main story Please verify you're not a robot by clicking the 
box. Invalid email address. Please re-enter. You must select a newsletter to subscribe to. Sign Up 
You will receive emails containing news content , updates and promotions from The New York Times. 
You may opt-out at any time. You agree to receive occasional updates and special offers for The New 
York Times's products and services. Thank you for subscribing. An error has occurred. Please try 
again later. View all New York Times newsletters.

Mr. Karzai said the agreement was "a good step toward the peace process" and called the peace 
agreement "a step solely that I didn't talk about."

On a day that saw conditions in rural Afghanistan triple-levels of temperature, the United States 
assessed a clear line of terms and amred along to Mr. Karzai's presidency. The decision was turned 
head over to the United States. Heads of Peace on Sunday dismissed reports that Mr. Sharif's forces 
had asked Mr. Karzai for talks to prevent his departure. A statement on Sunday said the Afghan 
government had "confirmed that Mr. Karzai's forces had been in force."

Advertisement Continue reading the main story

Mr. Karzai issued a statement hours after he amred the agreement, rejecting the fact that although 
the talks had ended, "extraordinary changes had been made for the recent developments."

The Afghan government's main said that, although the United States would resume talks before the 
parties signed, the agreement's effect dissolved in October and urged Mr. Karzai to stop giving 
discussions. But it also insisted that the United States would reach a bargaining gap between the 
parties before talks in Kabul and to compromise its efforts to pass a peace decree by the end of 
escalating disputes.

"We will never really allow both sides to negotiate," Mr. Karzai's spokesman said. "You don't 
negotiate a bad relationship when there's clear security."

Mr. Karzai and his troops were, in fact, supporters of both the United States and NATO.\textless{}\textbar{}endoftext\textbar{}\textgreater{}

\end{mdframed}

\paragraph{ELF-B example.}
8~NFE, $\Delta\mathrm{NLL}=-2.04$.  The \methodshort{} generation
recovers from token-level collapse into connected sentences about a sustained
topic.
\begin{mdframed}[style=samplebox]
\footnotesize
\textbf{Official}\par\smallskip
In conclusion conclusion,,,,:, conclusion,. maxim maxim. \textgreater{}\textgreater{} din., din, din, din., am, \textgreater{}\textgreater{},..,,.. 
\textgreater{}\textgreater{},., din din.,,,,,, din. din din,., din...,., din.. summarize.,,,,.

\par\medskip\textbf{\methodshort{}}\par\smallskip
In this article, a panel asked to determine the tank of the tank that is holding and the amount of 
tank that the tank is holding. Specifically the insights are determined by the size of the size of 
the tank tank, the location of the tank containers are holding and temperature, of the forced 
pressures of the tank held. The study is as described in the last part of the paper. In this 
article, again, calculations computes different features of tank storage, including 11 percent of 
the tank, half percent size included, 3/4 of the tank capacity, and the total amount of tank. If 
you want to use an increase in the tank, what is the total that you need? What can you use to 
determine how much tank you need? In this case, when you create a package so that once the maximum 
maximum capacity, the maximum can be estimated as 330 million USD or less because there are 20 
storage stations to deliver the total, you will need to add an additional amount of storage per 20 
storage containers to deliver the total. As  example, if you create a platform tank, it will 
deliver the total of about 20 tanks all in a day. If you want to create the tank in a specific 
location, you would also need to create a platform tank. You have to have the tank that stores the 
containers holding the tank. The platform platform also has a locked pit which dumps the garbage 
and the device to hold out tank to the tank. This way the whole tank must be in a tight underground 
pit of the water, and then the tank tank must be filled in water so that it can run properly. As 
long as there is not enough water to hold the tank, the decay is required to get the perfect tank 
of the tank. But the major advantage of a platform holding is you use it to store the storage and 
water and to transporte the data flows to foreign systems with a specific storage point. First and 
foremost, you ensure that the platform storage is exactly that is fit for the system you need. 
Platform holding is a staple of easily defined and easily integrated into your platform platform. 
Once you have a platform platform ready, you can create your own platform platform. Platform 
holding brings in some of the most advanced storage technologies available, and it includes 
countless different features and technical details that you can use. There are plenty of apps that 
go to you of how you want to to launch your own platform platform, and you can use them often when 
you need platform holding. The fundamentals of platform holding are one of the fundamental 
principles of platform holding. The features of platform holding are easily defined and include 
many different elements, including popular software, programable design tools, intelligent design 
plugins, color charts, color tables, graphs, data, and customized background effects. They can be 
specially customized and implemented in custom graphics, some that adapt custom design features for 
customized applications, such as charts and maps. They can also be specially customized with apps 
used to create user-based playlists. If you want an tank that is a platform platform, there are 
features to be used. For example, I remember my reading an avatar book. I remember I reading the 
Oldboy game, when I was 16, and I was screaming and the kid started to think and I was kill me, and 
then he pulled her on the floor and shot me down. I didn't want to see her, but I was 13, and I 
didn't want to look at that much. So I'm shut to sleep, and waste my life fighting and take the 
life, and let her go. I mean, she's really insane, and I can't stand her. And then I'm all out of 
love! So this is a great comic book in my life, and I have so much to share. I feel really wasted 
for reading, too. While there is still more to read, and I'm getting really excited, I'll make some 
some cool ideas that I find worth reading and reading. If you'd like to share your insight, tell us 
your thoughts and inspiration, or let us take part. The chance to explore it today, you we'll get 
to explore all of the beautiful locales where you'll want to hang together, walk the streets, walk 
the streets, and enjoy amazing games and competitions. Next, you'll have a new sport, our global 
video game, where you can raise your children, inspire family and friends, and show their faith and 
heart. In great respect to Andrew Berger, Tettarian anscientist, is a surgid and author of writing 
the Prophet Creid and War Naraous Abiql Abiq is a Jewish man who has been out of his own home in 
Iraq-occupied Islamic Arabia, for for the purpose of the killings of the Taliban while he 
undertakes the military operations in the nearby Iraq Arab militants forces.

\end{mdframed}

\endgroup

\end{document}